\documentclass[letterpaper, 10 pt]{IEEEtran}
\IEEEoverridecommandlockouts

\usepackage{url}
\usepackage[x11names]{xcolor}
\usepackage[colorlinks, bookmarksopen, bookmarksnumbered, citecolor=Red4, urlcolor=Turquoise4, linkcolor=Green4]{hyperref}

\usepackage{amssymb}
\usepackage{amsmath}
\usepackage{dsfont}
\usepackage{hyperref}
\usepackage{booktabs}
\usepackage{multirow}
\usepackage[caption=false]{subfig}
\usepackage[ruled, vlined]{algorithm2e}
\usepackage{natbib}
\usepackage{graphicx}
\usepackage{balance}

\DeclareMathOperator{\clip}{clip}
\DeclareMathOperator{\sign}{sign}
\DeclareMathOperator*{\argmax}{arg\,max}

\title{\Large \bf Combining RL and IL using a dynamic, performance-based modulation over learning signals and its application to local planning}

\author{Francisco Leiva$^{1}$ and Javier Ruiz-del-Solar$^{1}$%
\thanks{This work was supported by FONDECYT project 1201170, and ANID-PIA project AFB230001.}%
\thanks{$^1$Advanced Mining Technology Center (AMTC) and Department of Electrical Engineering, Universidad de Chile, Tupper 2007, Santiago, Chile.}%
\thanks{\tt\small{\{francisco.leiva, jruizd\}@ing.uchile.cl}}%
}

\begin{document}
\maketitle

\begin{abstract}

This paper proposes a method to combine reinforcement learning (RL) and imitation learning (IL) using a dynamic, performance-based modulation over learning signals. The proposed method combines RL and behavioral cloning (IL), or corrective feedback in the action space (interactive IL/IIL), by dynamically weighting the losses to be optimized, taking into account both the backpropagated gradients used to update the policy and the agent's estimated performance. In this manner, RL and IL/IIL losses are combined by equalizing their impact on the policy's updates, while simultaneously modulating said impact in such a way that imitation learning signals are prioritized at the beginning of the learning process, and as the agent's performance improves, the reinforcement learning signals  become progressively more relevant, allowing for a smooth transition from pure IL/IIL to pure RL. The proposed method is used to learn local planning policies for mobile robots, synthesizing IL/IIL signals online by means of a scripted policy. An extensive evaluation of the application of the proposed method to this task is performed in simulations, and it is empirically shown that it outperforms pure RL in terms of sample efficiency (achieving the same level of performance in the training environment utilizing approximately 4 times less experiences), while consistently producing local planning policies with better performance metrics (achieving an average success rate of 0.959 in an evaluation environment, outperforming pure RL by 12.5\% and pure IL by 13.9\%). Furthermore, the obtained local planning policies are successfully deployed in the real world without performing any major fine tuning. The proposed method can extend existing RL algorithms, and is applicable to other problems for which generating IL/IIL signals online is feasible.  A video summarizing some of the real world experiments that were conducted can be found in \url{https://youtu.be/mZlaXn9WGzw}.

\end{abstract}


\section{Introduction}
\label{sec:introduction}

The fast progress that machine learning has experienced in recent years has led to several learning frameworks being adopted to address important problems in robotics, such as navigation and its sub-tasks~\citep{xiao2022motion}. Reinforcement learning (RL) is one of said frameworks, and is characterized by addressing the problem of learning goal-directed behaviors from the interactions between an agent and its environment.

An appealing motif to train policies using RL, is that the agent being trained does not require explicit supervision: its behavior is modified during training so as to maximize a numerical reward signal \citep{sutton2018reinforcement}. This is specially convenient when addressing complex problems where hand-crafted solutions require extensive expert knowledge and tuning to work robustly.

Although several approaches to synthesize robot controllers using RL have been proposed in the past, there are still well-known issues that prevent their widespread adoption. In general, using RL to train controllers for robotic applications comes with a plethora of challenges~\citep{ibarz2021train}; for instance, the huge amount of agent-environment interactions usually required to obtain proficient policies, and the harm that some of these interactions may cause to the agent itself or to its environment if performed in the real world. To circumvent the above, a common practice in RL is to conduct the learning process in a simulation, and then to deploy the obtained controller in the real world. This practice, known as sim2real transfer \citep{zhao2020sim}, comes with the challenge of bridging the differences between the simulation and the real world to be successfully applied.

Alternative methods for learning control policies, such as behavioral cloning (BC) under the imitation learning (IL) framework, tend to be much more sample efficient than RL-based formulations, however, controllers obtained using BC heavily rely on the availability and quality of the data set they are trained on, often failing to behave properly when facing previously unseen situations, and being unable to surpass the performance level of the expert that is being imitated.

Given the advantages and limitations of both RL and IL, we propose a general approach to harness the sample efficiency of IL-based methods to accelerate the reinforcement learning of robot control policies. We present a method that incorporates learning signals coming from imitation or interactive imitation learning (IIL) into off-policy RL algorithms by modulating them based on an estimation of the agent's performance during training.  

The proposed method takes into account the problem of meaningfully combining RL and IL losses (given their different nature), so as to equalize their impact on the policy's parameter updates (according to back-propagated gradients), while modulating said impact by means of the agent's performance estimate, thus, allowing for a smooth transition from pure IL/IIL to pure RL. Moreover, these two mechanisms do not introduce additional parameters to those already required to formulate a given problem under the RL framework.

To validate the proposed method, we address the reinforcement learning of local planning policies for robotic navigation, i.e., the use of RL to obtain controllers that would allow a mobile agent to reach a local target destination using local information. In the literature, this problem is often referred to as ``mapless navigation'', since the trained agents are also characterized by not requiring a map of the environment in which they are deployed to reach a local navigation target. The required IL and IIL signals are synthesized online using a scripted policy based on the pure pursuit path tracking algorithm \citep{coulter1992implementation}, so no prior demonstration data nor online human interaction is required to train the policies.

For the real world application of the proposed method, we present a streamlined pipeline for the sim2real transfer of local planning policies in a real skid-steered robot, characterized for being hard to simulate with high fidelity.

We assess the proposed method taking into account its sample efficiency, and the obtained policies' performance and generalization capabilities empirically. 

Thus, the main contributions derived from this work are the following:
\begin{itemize}
    
    \item A method that allows combining RL and IL by dynamically weighting the losses to be optimized, taking into account their impact on the policy's updates and modulating said impact based on the agent's estimated performance, thus, inducing a smooth transition from pure IL/IIL, to pure RL.
    
    \item The application of the proposed method to learn local planning policies, using imitation learning signals synthesized online using a scripted policy.

    \item A comparison of the proposed method with pure RL and pure IL/IIL in terms of sample efficiency, and the performance of the obtained policies for the local planning problem. 
    
    \item A general pipeline to train and deploy local planning policies using the proposed method, addressing the sim2real transfer problem.
    
\end{itemize}

\section{Related work}
\label{sec:related_work}

In what follows, we present a brief literature overview of works that address the local planning problem (or related tasks), using reinforcement learning (RL), imitation learning (IL), and interactive imitation learning (IIL). Moreover, we also discuss several works that propose hybrid learning regimes by combining different learning frameworks, namely, IL/IIL and RL.

\subsection{Reinforcement Learning}

Several RL-based local planners for robotic navigation have been proposed in the recent years \citep{zhu2021deep}. These works cover a wide variety of formulations for the problem. For instance, depending on what the agent observes, there are local planners that rely on visual information \citep{zhu2017target,lobos2018visual}, fixed range measurements \citep{tai2017virtual, francis2020long}, processed range measurements \citep{zhang2022ipaprec}, or point clouds \citep{leiva2020robust,lobos2020point,zhang2021learn}. In addition, depending on how the agent's actions are modelled, there are works that focus on discrete or continuous control.

Most of these local planners are trained in simulations and then deployed in the real-world (sim2real), although often not directly due to the simulation to reality gap \citep{kadian2020sim2real}. The training of these policies is conducted in simulations mainly because of the large number of agent-environment interactions required for the learning process, and the risk that performing some of these interactions in the real world might entail.

\subsection{Imitation Learning}

Imitation learning methods can be roughly categorized in inverse reinforcement learning (IRL) and behavioral cloning (BC) methods \citep{osa2018algorithmic,zheng2022imitation}. IRL focuses on learning an unknown reward function from demonstrations, and then using this reward to obtain a policy (e.g., using RL). BC, on the other hand, addresses the problem of learning a mapping between observations and actions from labeled data, without estimating (nor requiring) a reward function in the process. In this regard, the method we present relates to BC-based local planners, such as those proposed by \cite{gao2017intention} and \cite{pfeiffer2017perception}, which are obtained by learning in a supervised manner, using a data set of expert demonstrations.

An issue that commonly arises when using standard BC, is the compounding of errors that occurs due to a mismatch between the training and testing data sets \citep{ross2010efficient}. To alleviate this issue, data observed by the learning agent can be labeled and added to the training data set, as proposed by \cite{ross2011reduction,ross2013learning} in the DAgger (Dataset Aggregation) algorithm. 

\subsection{Interactive Imitation Learning}

Another approach to train policies is to use interactive imitation learning (IIL), where a teacher provides feedback to the learning agent to improve its behavior \citep{celemin2022interactive}. While there are many modalities for teacher-agent interaction, in this work we focus on relative corrections in the action space, that is, in interactions where the teacher provides feedback that aims to increase or decrease the magnitude of the actions performed by the agent. While this type of interaction requires knowing how changes in the agent's actions affect its performance for a given task, it relieves the teacher from knowing exactly how to act~\citep{celemin2022interactive}. This interaction modality has been previously explored in the Advice-Operator Improvement (\mbox{A-OPI}) algorithm \citep{argall2008learning}, as well as in the COrrective Advice Communicated by Humans (COACH) framework \citep{celemin2019interactive}. Moreover, \cite{perez2019continuous,perez2020interactive} addressed car-racing like tasks using COACH variants designed to parameterize policies as deep neural networks.

\subsection{Hybrid methods}

Due to the advantages and disadvantages that IL, IIL and RL possess, there have been efforts in developing hybrid methodologies based on said frameworks. An idea underlying the combination of RL and IL/IIL is that of ``side-stepping'' exploration~\citep{ibarz2021train}, that is, to use IL/IIL to guide the agent into experiencing informative interactions with its environment, thus, aiding its learning process. As examples of the above, Pfeiffer et al. proposed accelerating the reinforcement learning of local planning policies by pre-training them using BC~\citep{pfeiffer2018reinforced}, while the incorporation of demonstrations directly into RL algorithms was explored, for instance, in the DQfD~\citep{hester2018deep}, the DDPGfD~\citep{vecerik2017leveraging}, and the DAPG~\citep{rajeswaran2018learning} algorithms. 

The combination of IIL and RL has also been studied in prior works. Celemin et al., for instance, explored harnessing human corrective feedback within RL \citep{celemin2019fast, celemin2019reinforcement} by interleaving policy updates based on COACH during the execution of rollouts, and policy search updates.

The way in which different learning signals should be combined, and how this combination should evolve, has also been studied. \cite{zhu2018reinforcement}, for instance, utilized a fixed factor to augment the RL objective using an IL reward component based on the GAIL algorithm \citep{ho2016generative}. Similarly, \cite{zolna2019reinforced} also augmented the RL objective using IL rewards which take inspiration on the GAIL algorithm (although they only depend on state observations), however, these IL rewards are only taken into account given the value of a binary random variable that depends on the policy's estimated performance. In this manner, there are interleaved policy updates using the RL objective augmented with IL, and the standard RL objective. As the learning process progresses and the agent's performance increases, pure RL updates become more likely, thus, avoiding the potential limitations of imitating an expert.

\cite{schmitt2018kickstarting} proposed combining the RL objective with an IL loss term modulated by a variable factor to accelerate the learning process of agents. The modulation of the IL loss term was performed using Population Based Training, but constant and linear schedules to vary the modulation were also explored.

The Cycle-of-Learning (CoL) framework, proposed by~\cite{waytowich2018cycle}, presents a similar idea to leverage different human interaction modalities to aid the learning process of autonomous systems. These different interaction modalities, namely, demonstrations, interventions and evaluations, are integrated and progressively switched so as to get to a pure RL regime. Under the CoL framework, \cite{goecks2019efficiently} explored how to leverage demonstrations and interventions, and demonstrations and reinforcement learning to autonomously land a quadrotor \citep{goecks2020integrating}.

In this study, we modulate the combination of RL and IL/IIL signals using the agent's estimated performance to smoothly transition from pure IL/IIL to pure RL. However, unlike prior works, we take into account the backpropagated gradients utilized to update the policy, thus, addressing the problem of effectively combining learning signals of different nature, while purposefully changing their relative importance as the training process progresses. To balance the RL and IL/IIL losses, a strategy similar to that proposed in the GradNorm algorithm \citep{chen2018gradnorm} is proposed, and this balancing strategy is carried out alongside the performance-based modulation of RL and IL/IIL signals. Furthermore, we show that our method can be utilized either using IL signals based on BC or IIL signals based on relative corrections in the action space.

\section{Combining RL and IL signals using a performance-based modulation strategy}
\label{sec:proposed_approach}

\subsection{Problem formulation}
\label{subsec:problem_formulation}

Let us consider an agent interacting with the environment. This agent-environment interaction is modeled as a Partially Observable Markov Decision Process (POMDP) defined by a set of states $\mathcal{S}$, a set of actions $\mathcal{A}$, a reward function $\mathcal{R}\left(s,a\right)$, a state transition function $\mathcal{T}\left(s,a,s'\right)=p\left(s'|s,a\right)$, a set of observations $\Omega$, an observation function $\mathcal{O}\left(s',a,o\right)=p\left(o|s', a\right)$, and a discount factor $\gamma \in \left[0,1\right)$. At each time step $t$ the agent is in some state $s_t$, observes $o_t$, executes an action $a_t$ according to its policy $\pi\left(a_t|o_t\right)$, receives a scalar reward $r_t$, and experiences the transition to a new state, $s_{t+1}$. 

The goal is to learn a policy that maximizes the expected discounted return that the agent would get in its interaction with the environment, that is, to maximize Eq.~\eqref{eq:rl_objective}.
\begin{equation}
    J_\text{RL}\left(\pi\right)=\mathbb{E}_{a_t\sim \pi\left(a_t|o_t\right)}\left[\sum^T_{t=1}\gamma^{t-1}r_t\right]
    \label{eq:rl_objective}
\end{equation}

We assume that the agent's policy $\pi$ is parameterized by parameters $\phi$, thus, we denote it by $\pi_\phi$. With the above, the goal becomes finding the parameters $\phi^{*}$ that satisfy Eq.~\eqref{eq:rl_objective_2}.

\begin{equation}
    \phi^{*} = \argmax_{\phi} J_\text{RL}(\pi_{\phi})
    \label{eq:rl_objective_2}
\end{equation}

\subsection{Synthesizing imitation learning signals}
\label{subsec:supervised_learning_signals_synthesis}

In the standard RL framework, learning a policy is done by relying on the experiences generated due to the interactions between the agent and its environment. However, to accelerate the agent's learning process, additional signals may be generated and utilized during training. 

If expert demonstrations or feedback regarding the agent's behavior is available online, two straightforward approaches may be followed to synthesize such learning signals.

\subsubsection{Imitation Learning}
\label{subsubsec:il_approach}

If the actions that the agent should perform can be queried to an expert, then a standard behavioral cloning approach can be followed. Let us denote the expert policy as $\pi_\text{expert}(a|o_\text{expert})$, where $o_\text{expert}$ potentially includes privileged information. Then, by collecting a data set of observation-actions pairs generated by this expert, $\mathcal{D}= \left\{\left(o_i, a_i\right)\right\}_{i=1}^N$, we can try to imitate its behavior by minimizing the objective defined by Eq.~\eqref{eq:bc_loss}, where $\mathcal{L}$ is an ad-hoc cost function.
\begin{equation}
\label{eq:bc_loss}
    J_\text{BC}(\pi_\phi) = \frac{1}{N}\sum_{i=1}^N \mathcal{L}\left(\pi_\phi(a|o_i), a_i\right)
\end{equation}

It is important to remark that the observations in $\mathcal{D}$ are those that the learning agent would obtain, and not those received by the expert policy.

\subsubsection{Interactive Imitation Learning}
\label{subsubsec:iil_approach}

Let us suppose we do not have access to an expert demonstrator, but to an expert that only provides feedback on the actions performed by the agent. In this scenario, by adopting the COrrective Advice Communicated by Humans (COACH) framework~\citep{celemin2019interactive}, we can also generate a feedback signal to shape the agent's policy. Under the COACH framework, said feedback is discrete, and aims to induce a magnitude change in the actions executed by the agent. For a given action $a_t$, let us denote the expert's feedback as $h_t \in \left\{-1, 0, 1\right\}^{n_a}$, where $n_a$ is the actions' dimension. The objective of this feedback is to correct the executed action's components depending on whether or not their magnitudes were adequate according to the expert. Thus, given a fixed parameter $e>0$, the corrected action $a_t + e\cdot h_t$ is utilized as a label for the observation-action mapping that is being learned. Since $h_t$ is only valid for $t$, in an extension to COACH, called \mbox{D-COACH}~\citep{perez2020interactive}, it is proposed to use a small ring buffer to store the action labels and their corresponding observations to allow batch-training of policies. Therefore, given a data set $\mathcal{D}=\left\{\left(o_i, a_i, h_i\right)\right\}_{i=1}^N$ and a suitable cost function $\mathcal{L}$, we can shape the policy's actions by minimizing the objective defined by Eq.~\eqref{eq:coach_loss}.
\begin{equation}
\label{eq:coach_loss}
    J_\text{COACH}(\pi_\phi) = \frac{1}{N}\sum_{i=1}^N \mathcal{L}\left(\pi_\phi(a|o_i), a_i + e\cdot h_i\right)
\end{equation}

\subsection{Combining RL and IL using performance-modulated learning (PModL)}
\label{subsec:iconcorporating_il_to_rl}

To incorporate the learning signals described in Section~\ref{subsec:supervised_learning_signals_synthesis} into the RL framework, we propose utilizing a modulation strategy that depends on the policy's performance during training. To measure the policy's performance, we compute an estimate of its success rate as the ratio of successful episodes over the the last $K_\text{SR}$ episodes. Denoting this estimate as $z\in [0,1]$, the policy's parameters are updated by minimizing the ``performance-modulated learning'' (PModL) objective defined by Eq.~\eqref{eq:pmodil_loss}, where $J_\text{RL}(\pi_\phi)$ is the policy's RL objective, $J_\text{IL}(\pi_\phi)$ is the BC or COACH objective (see Eqs.~\eqref{eq:bc_loss} and~\eqref{eq:coach_loss}), and $\lambda > 0$.
\begin{equation}
    \label{eq:pmodil_loss}
    J_{\text{PModL}}(\pi_\phi) = z J_\text{RL}(\pi_\phi) + \lambda (1-z) J_\text{IL}(\pi_\phi)
\end{equation}

In Eq.~\eqref{eq:pmodil_loss}, the performance estimate $z$ modulates the relative importance of $J_\text{RL}(\pi_\phi)$ and $J_\text{IL}(\pi_\phi)$ as the learning process progresses, and $\lambda$ serves as a balancing factor to equalize the impact of these terms, given their different nature. 

While $z$ can be easily computed during training, since $J_\text{RL}(\pi_\phi)$ depends on the reward function (which varies across tasks and design decisions for the problem formulation), $\lambda$ requires tuning. Although adjusting $\lambda$ via grid search is a viable option, this would be computationally expensive. Instead, we propose dynamically adjusting $\lambda$ so as to balance the gradient magnitudes related to the RL and IL/IIL losses, thus, making the computation of $J_{\text{PModL}}(\pi_\phi)$ not introduce additional parameters.

\subsubsection{Computing $z$}
As stated previously, the policy's performance estimate is set to be the ratio of successful episodes over the last $K_\text{SR}$ episodes. To compute $z$, we consider a circular buffer $\mathcal{Z}$ of size $K_\text{SR}$, initialized with all of its elements set to zero.

Whenever an episode ends, $\text{success}(episode) \in \{0, 1\} $ is saved in $\mathcal{Z}$, where $\text{success}(episode)$ indicates whether an episode was successful or not, and $episode$ may be arbitrary information available in the episode itself (e.g., states, actions, instantaneous rewards, etc., that would be needed to determine if the episode was successful). 

With the above, $z$ is simply computed as the average value of the elements contained in $\mathcal{Z}$. Notice that initializing $\mathcal{Z}$ with all its elements set to zero, prevents an overestimation of the policy's performance at the beginning of the training process.

\subsubsection{Computing $\lambda$}
\label{subsubsec:computing_lambda}

To compute $\lambda$, we propose following a strategy similar to that presented in the GradNorm algorithm \citep{chen2018gradnorm} for multitask loss balancing. 

Denoting the parameters of the last layer of the policy parameterization by $W$, and the $L_2$ norm of the $J_\text{RL}(\pi_\phi)$ and $\lambda J_\text{IL}(\pi_\phi)$ gradients with respect to $W$ by $G^{\text{RL}}_W$ and ${}^{\lambda}G^{\text{IL}}_W$, respectively (see Eqs.~\eqref{eq:w_rl_grads} and~\eqref{eq:w_lambda_il_grads}, we define the loss function utilized to adjust $\lambda$ by Eq.~\eqref{eq:lambda_loss}.
\begin{align}
    G^{\text{RL}}_W &= \left\| \nabla_W J_\text{RL}(\pi_\phi)\right\|_2 \label{eq:w_rl_grads}\\
    {}^{\lambda}G^{\text{IL}}_W &= \left\| \nabla_W \lambda J_\text{IL}(\pi_\phi)\right\|_2 \label{eq:w_lambda_il_grads} \\
    J_\lambda(\lambda) &= \left|{}^{\lambda}G^{\text{IL}}_W - G^{\text{RL}}_W \right|_1 \label{eq:lambda_loss}
\end{align}

The idea is to adjust $\lambda$ by minimizing Eq.~\eqref{eq:lambda_loss} via gradient descent, where $G^{\text{RL}}_W$ should be interpreted as a constant, and $J_\lambda(\lambda)$ is differentiated with respect to $\lambda$. To avoid the scenario in which $\lambda$ goes to zero, after each update it is set to be at least equal to $\lambda_\text{min} > 0$, that is, $\lambda \leftarrow \max\{\lambda, \lambda_\text{min}\}$. To avoid coupling $\lambda$ with the policy's learning rate, we suggest setting $\lambda_\text{min}=1$.

Intuitively, $\lambda$ will vary dynamically to balance the backpropagated gradient magnitudes associated to RL and IL/IIL, making the performance estimate $z$ responsible for the smooth transition from pure IL/IIL ($z=0$) to pure RL (in the ideal case in which $z=1$), despite the different nature of $J_\text{RL}(\pi_\phi)$ and $J_\text{IL}(\pi_\phi)$.

Given the above strategy, $\lambda$ should be updated every time $J_{\text{PModL}}(\pi_\phi)$ is utilized to update the policy's parameters. In a standard training procedure, that means that every time a batch of data is utilized to perform an update, $\lambda$ should be modified prior to the computation of $J_{\text{PModL}}(\pi_\phi)$.

\subsubsection{Integrating PModL into existing RL algorithms} 
Note that given that PModL specifies a loss that combines RL and IL to adjust the policy's parameters, it is suitable to extend a wide variety of pre-existing RL algorithms. 

An algorithm that incorporates the PModL extension is described in Section~\ref{subsec:practical_implementation}, where Deep Deterministic Policy Gradient \citep{lillicrap2015continuous} is utilized as the base RL algorithm. In this DDPG extension, denoted as ``DDPG+PModL'', either IL or IIL signals can be selected as $J_\text{IL}(\pi_\phi)$, and are generated online utilizing a scripted policy.

\section{Addressing the local planning problem with PModL}
\label{sec:adressing_lp}

\subsection{Problem formulation}
\label{subsec:problem_formulation_lp}

As stated in Section~\ref{sec:introduction}, the local planning problem consists of getting a mobile agent to reach a local navigation target using local information of the environment with which it interacts.

\begin{figure}
    \centering
    \includegraphics[width=0.8\linewidth]{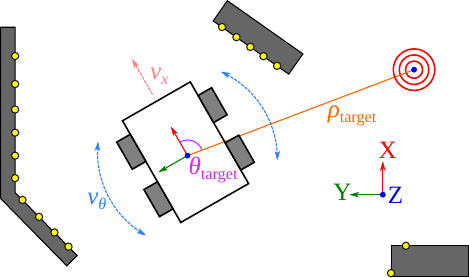}
    \caption{Diagram of the problem addressed in this work. The velocity commands and odometry-based estimations are referenced to the robot's local frame. The linear velocity of the robot, $v_x$, goes along the $X$-axis, and its angular velocity, $v_\theta$, around the $Z$-axis. The navigation target is defined in polar coordinates as $(\rho_\text{target}, \theta_\text{target})$, also with respect to the robot's local frame. The 2D range measurements are represented as yellow dots.}
    \label{fig:problem_formulation}
\end{figure}

Fig.~\ref{fig:problem_formulation} illustrates the addressed problem. We will assume that the mobile robot to be controlled is nonholonomic, responds to continuous linear and angular velocity commands, and has access to odometry-based velocity estimations. Furthermore, we will assume that the terrain in which the robot navigates only requires 2D perception to be properly characterized, and that the robot is equipped with at least one sensor that provides 2D range measurements. Finally, we will also assume that although the environment is unknown, and the agent does not have access to a map of any kind, it does have access to a localization system that enables it to keep track of its pose with respect to a local navigation target.

\subsection{Local planning as a POMDP}

Given the formulation presented in Section~\ref{subsec:problem_formulation}, the agent-environment interaction is modeled as a POMDP, therefore, in what follows, the agent's observations, actions, reward function, and episodic settings are defined.

\subsubsection{Observations}
\label{subsec:observations}
 
Following the approach proposed by \cite{leiva2020robust}, the agent's observations are defined according to Eq.~\eqref{eq:observations}, where $o^t_\text{pcl}$ is a point cloud that encapsulates the information provided by 2D LiDARs, $o^t_\text{odom}$ contains odometry-based speed estimations of the robot, and $o^t_\text{target}$ is the local target that has to be reached.
\begin{equation}
\label{eq:observations}
    o_t = \left(o^t_{\text{pcl}}, o^t_\text{odom}, o^t_\text{target}\right)
\end{equation}
As in the work of \cite{leiva2020robust}, $o^t_\text{pcl}$ is constructed using a fixed-size vector containing the $n$ range measurements provided by a 2D LiDAR, $o^t_{\text{range}}=\left(\rho^t_1,..., \rho^t_n\right)$, where each measurement is such that, for $i \in \left\{1,..., n\right\}$, $\rho^t_i \in \left[\rho_\text{min}, \rho_\text{max}\right]\cup\left\{\underline{\rho}, \overline{\rho}\right\}$, where $\rho_\text{min}$ and $\rho_\text{max}$ are the minimum and maximum range values, and $\underline{\rho}$ and $\overline{\rho}$ are codifications for range measurements that are below $\rho_\text{min}$ or above $\rho_\text{max}$, respectively. As each range measurement $\rho^t_i$ has an associated angle $\theta^t_i$, then $o^t_\text{pcl}$ is defined according to Eq.~\eqref{eq:o_pcl}, where $k_t\leq n$ is the number of in-range measurements for time step $t$.
\begin{equation}
\label{eq:o_pcl}
    o^t_\text{pcl} = \left\{\left(\frac{\rho^t_j}{\rho_\text{max}}\cos\left(\theta^t_j\right), \frac{\rho^t_j}{\rho_\text{max}}\sin\left(\theta^t_j\right)\right)\right\}_{j=1}^{k_t\leq n}
\end{equation}

The component $o^t_\text{odom}$, on the other hand, is defined as a vector containing the normalized odometry-based linear and angular velocity estimations of the robot. Denoting $v^t_{\text{odom}-x}$ and $v^t_{\text{odom}-\theta}$ as the unnormalized odometry-based linear and angular velocity estimations, respectively, and $v_x^{\text{max}}$ and $v_\theta^{\text{max}}$ as their maximum values, then $o^t_\text{odom}$ is defined by Eq.~\eqref{eq:o_odom}. 

Notice that in an ideal case, the restrictions placed on the agent's actions (see Section~\ref{subsec:actions}) result on $0 \leq \tilde{v}^t_{\text{odom}-x} \leq 1$ and $-1 \leq \tilde{v}^t_{\text{odom}-\theta} \leq 1$.
\begin{equation}
\label{eq:o_odom}
    o^t_\text{odom}=\left(\frac{v^t_{\text{odom}-x}}{v_x^{\text{max}}}, \frac{v^t_{\text{odom}-\theta}}{v_\theta^{\text{max}}}\right) = \left(\tilde{v}^t_{\text{odom}-x}, \tilde{v}^t_{\text{odom}-\theta}\right)
\end{equation}

Finally, $o^t_\text{target}$ is defined as a vector that contains the normalized polar coordinates of the local navigation target. By setting a maximum distance for the local targets' generation, $\rho^\text{max}_\text{target}$, and considering that the angle to the target, $\theta^t_\text{target}$, is lower bounded by $-\pi$ and upper bounded by $\pi$, then $o^t_\text{target}$ is given by Eq.~\eqref{eq:o_target}.
\begin{equation}
\label{eq:o_target}
    o^t_\text{target}= \left(\frac{\min\left\{\rho^t_\text{target}, \rho^\text{max}_\text{target}\right\}}{\rho^\text{max}_\text{target}}, \frac{\theta^t_\text{target}}{\pi}\right)= \left(\tilde{\rho}^t_\text{target}, \tilde{\theta}^t_\text{target}\right)
\end{equation}

\subsubsection{Actions}
\label{subsec:actions}

As the mobile robot we are considering responds to continuous velocity commands (see Section~\ref{subsec:problem_formulation_lp}), the actions are defined as a two-dimensional vector containing these commands, that is, $a_t=\left(v^t_{\text{cmd}-x}, v^t_{\text{cmd}-\theta}\right)$, where $v^t_{\text{cmd}-x}\in\left[v^\text{min}_{\text{cmd}-x}, v^\text{max}_{\text{cmd}-x}\right]$ is a linear velocity command, and $v^t_{\text{cmd}-\theta}\in\left[v^\text{min}_{\text{cmd}-\theta}, v^\text{max}_{\text{cmd}-\theta}\right]$ is an angular velocity command. Furthermore, we set $v^\text{min}_{\text{cmd}-x} = 0$, and $-v^\text{min}_{\text{cmd}-\theta} = v^\text{max}_{\text{cmd}-\theta}$.

\subsubsection{Reward function}
\label{subsec:reward_function}

Following a similar approach as that proposed by~\cite{leiva2020robust}, the reward function is defined by Eq.~\eqref{eq:reward_function}, where $\rho_\text{thresh}$ corresponds to a threshold distance utilized to decide whether the mobile robot has reached the navigation target or not, whilst $r_\text{success}$ and $r_\text{collision}$ are fixed scalar values.
\begin{equation}
\label{eq:reward_function}
    r_t = \begin{cases}
    r^t_\text{speed} + r^t_\text{target} +  r^t_\text{fov} + r^t_\text{danger} & \text{if $\rho^t_\text{target} \geq \rho_\text{thresh}$},\\
    r_\text{success} & \text{if $\rho^t_\text{target} < \rho_\text{thresh}$}, \\
    r_\text{collision} & \text{if the agent collides}.
    \end{cases}
\end{equation}

The terms $r^t_\text{speed}$ and $r^t_\text{target}$ encourage the agent to reach the navigation target (Eqs.~\eqref{eq:r_speed} and~\eqref{eq:r_target}). The term $r^t_\text{fov}$ rewards the agent if the navigation target lies within a $2\theta_\text{fov}$ radians field of view (FoV) projected from the robot's local frame (Eq.~\eqref{eq:r_fov}). Finally, $r^t_\text{danger}$ penalizes the agent if it dangerously approaches an obstacle, that is, if the minimum value of any of its in-range LiDAR measurements is less than a predefined safety distance, $\rho_\text{danger-thresh}$ (Eq.~\eqref{eq:r_danger}). 

The variables $K_\text{speed}$, $K_\text{target}$, $K'_\text{target}$, $K_\text{fov}$, $K'_\text{fov}$ and $K_\text{danger}$, used in the definition of these terms, correspond to fixed scalar values.
\begin{align}
    r^t_\text{speed} &= K_\text{speed}\tilde{v}^t_{\text{odom}-x}\cos\left(\theta^t_\text{target}\right)\label{eq:r_speed}\\
    r^t_\text{target} &= \mathds{1}_{\left\{\rho^t_\text{target} < \rho^{t-1}_\text{target}\right\}}K_{\text{target}} -K'_\text{target} \label{eq:r_target} \\
    r^t_\text{fov} &= \mathds{1}_{\left\{|\theta^t_\text{target}| > \theta_\text{fov} \right\}}\left(K_\text{fov}\cos\left(\theta^t_\text{target}\right) - K'_\text{fov}\right)\label{eq:r_fov}\\
    r^t_\text{danger} &= \mathds{1}_{\left\{\min\limits_{j=1,...,k_t}\left\{\rho^t_j\right\} < \rho_\text{danger-thresh}\right\}}K_\text{danger} \label{eq:r_danger}
\end{align}

\subsubsection{Episodic settings}
\label{subsec:episodic_settings}

The local planning problem, as formulated in this work, is episodic: an episode starts with the mobile robot in a randomly selected valid initial pose, and a random navigation target is then chosen such that $\rho^t_\text{target} \geq \rho_\text{thresh}$; on the other hand, an episode ends when the robot collides, or when it reaches the navigation target, that is, when $\rho^t_\text{target} < \rho_\text{thresh}$.

The initial pose for the robot is selected such that, given the execution of any action, a collision would not be possible after a single time step. Furthermore, to avoid episodes in which reaching the navigation target would only require short, trivial action sequences, the navigation target is selected such that $\rho^t_\text{target} \geq \rho_\text{init-thresh}> \rho_\text{thresh}$, that is, the selected navigation target has to be at a minimum distance of $\rho_\text{init-thresh}$ from the initial robot's position.

\subsection{Synthesizing IL signals for the local planning problem}
\label{subsec:synthesizing_il_lp}

As stated in Section~\ref{subsec:supervised_learning_signals_synthesis}, the agent's learning process may be accelerated if online expert demonstrations or feedback are available. 

For the local planning problem, let us consider a collision-free path to the navigation target, $\mathcal{C}_t$, generated by a global planner which has access to a metric map of the environment. If $\mathcal{C}_t$ exists for every time step, then a strategy to reach the navigation target would be to select actions $a_t$ that allow the agent to get close to a sampled nearby position within the path (that is, to a way point).

Given the existence of a collision-free path, $\mathcal{C}_t$, that connects the current agent position with the local navigation target, we consider a slightly modified pure pursuit path tracker \citep{coulter1992implementation} as an expert demonstrator. This path tracker provides control actions to get near a local way point $\left(x^t_\text{wp}, y^t_\text{wp}\right)$ in $\mathcal{C}_t$ that is one ``look-ahead'' distance, $\rho_{\text{look-ahead}}$, from the mobile robot. For a differential, nonholonomic robot with the restrictions $v^\text{min}_{\text{cmd}-x} = 0$, and $-v^\text{min}_{\text{cmd}-\theta} = v^\text{max}_{\text{cmd}-\theta}$ (see Section~\ref{subsec:actions}), the path tracker provides actions according to Eqs.~\eqref{eq:path_tracker_vel_x} and~\eqref{eq:path_tracker_vel_theta}\footnote{The function $\clip\{x, a, b\}$ is defined as $\min\{\max\{x, a\}, b\}$.}, where $K_x > 0$ is a fixed gain, $\theta^t_\text{wp}$ is the angle between the robot's local frame and the way point, $\theta^{+}_\text{wp}>0$ is a threshold that forces a zero linear speed unless an alignment requirement, $\left|\theta^t_\text{wp}\right| < \theta^{+}_\text{wp}$, is met, and $\theta^{-}_\text{wp}>0$ is another threshold set to prevent low angular speeds when the robot backward faces the target way point.
\begin{align}
    v^t_{\text{pp}-x} &= \clip\left\{\mathds{1}_{\left\{|\theta^t_\text{wp}| < \theta^{+}_\text{wp} \right\}}K_x x^t_{\text{wp}}, v^{\text{min}}_{\text{cmd}-x}, v^{\text{max}}_{\text{cmd}-x}\right\}\label{eq:path_tracker_vel_x}\\
    v^t_{\text{pp}-\theta} &=  \begin{cases}
        \clip\Big\{2\frac{y^t_\text{wp}}{{\rho_{\text{look-ahead}}}^2}, \\ \qquad \qquad \quad v^{\text{min}}_{\text{cmd}-\theta}, v^{\text{max}}_{\text{cmd}-\theta}\Big\} & \text{if } |\theta^t_\text{wp}| < \theta^{-}_\text{wp},\\
        \mathds{1}_{\left\{\theta^t_\text{wp} \geq 0 \right\}}v^{\text{max}}_{\text{cmd}-\theta} \\ \qquad \qquad + \mathds{1}_{\left\{\theta^t_\text{wp} < 0 \right\}}v^{\text{min}}_{\text{cmd}-\theta} & \text{otherwise.}
    \end{cases} \label{eq:path_tracker_vel_theta}
\end{align}

Using this path tracker as an expert policy $\pi_\text{expert}$, it is possible to collect observation-action pairs and follow the IL approach described in Section~\ref{subsubsec:il_approach}. 

For the IIL approach under the COACH framework (see Section~\ref{subsubsec:iil_approach}), we can also synthesize a feedback signal, $h_t = (h^t_{\text{cmd}-x}, h^t_{\text{cmd}-\theta})$. Defining $\Delta v^t_{\text{cmd}-x} = v^t_{\text{pp}-x} - v^t_{\text{cmd}-x}$ and $\Delta v^t_{\text{cmd}-\theta} = v^t_{\text{pp}-\theta} - v^t_{\text{cmd}-\theta}$, $h_t$ can be obtained using Eqs.~\eqref{eq:ht_x} and \eqref{eq:ht_theta}, where $\Delta v^{\text{thresh}}_{\text{cmd}-x}$ and $\Delta v^{\text{thresh}}_{\text{cmd}-\theta}$ are fixed thresholds.
\begin{align}
    h^t_{\text{cmd}-x} &= \mathds{1}_{\left\{\left|\Delta v^t_{\text{cmd}-x}\right| > \Delta v^{\text{thresh}}_{\text{cmd}-x} \right\}} \sign\left(\Delta v^t_{\text{cmd}-x}\right) \label{eq:ht_x}\\
    h^t_{\text{cmd}-\theta} &= \mathds{1}_{\left\{\left|\Delta v^t_{\text{cmd}-\theta}\right| > \Delta v^{\text{thresh}}_{\text{cmd}-\theta} \right\}} \sign\left(\Delta v^t_{\text{cmd}-\theta}\right) \label{eq:ht_theta}
\end{align}

\begin{figure}
    \centering
    \includegraphics[width=0.9\linewidth]{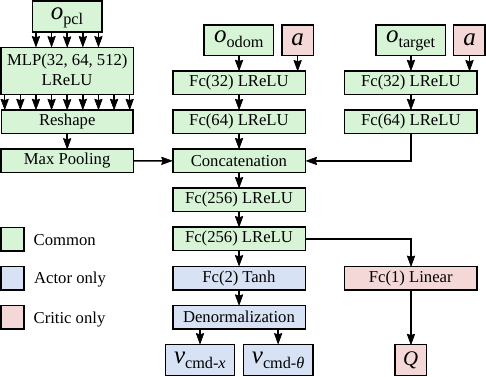}
    \caption{Diagram of the actor and critic architectures. The layers are described by the operation they perform or by a ``Type(Parameters) Activation Function'' notation. ``Fc'' stands for fully connected, and its parameter to the number of units it has. ``MLP'' stands for multilayer perceptron, and each of its parameters corresponds to the number of units of each fully connected layer that conforms the MLP. Finally, ``LReLU'' stands for Leaky ReLU and ``Tanh'' for hyperbolic tangent.}
    \label{fig:policy_architecture}
\end{figure}

\subsection{RL algorithm and policy parameterization}

Given the agent's observations (see Section~\ref{subsec:observations}), we parameterize its policy as a multimodal neural network. While many deep RL algorithms would be suitable to learn such policy, we use the Deep Deterministic Policy Gradient (DDPG) algorithm~\citep{lillicrap2015continuous}, as it has been widely adopted for solving the local planning problem previously, and it is already sample efficient due to its off-policy nature. Since DDPG is an actor-critic algorithm, two independent neural networks, $\pi_\phi$ and $Q_\psi$, are utilized to parameterize the policy (actor) and the Q-function (critic), respectively.

The actor and critic architectures are based on the multimodal ``PCL'' models proposed by~\cite{leiva2020robust}, and are shown in Fig.~\ref{fig:policy_architecture}. These neural networks are able to take $o_\text{pcl}$ as input by using a module akin to the feature extractor proposed for the PointNet architecture \citep{qi2017pointnet}. The rest of the observation's components, $o_\text{odom}$ and $o_\text{target}$, are processed independently by two fully connected layers, and the resulting intermediate representations are concatenated with the PointNet-based feature extractor's output. The resulting representation is then fed to a sequence of fully connected layers whose output is mapped to the action's numerical interval, for the actor, or directly utilized as a Q-value estimate, for the critic. These neural networks, although illustrated in a single diagram in Fig.~\ref{fig:policy_architecture}, do not share weights.

\subsubsection{Practical implementation}
\label{subsec:practical_implementation}
The combination of the PModL objective and the DDPG algorithm (DDPG+PModL) is shown in Algorithm~\ref{algo:ddpg+pmodl}.

\setlength{\algomargin}{5pt}
\begin{algorithm}

\small
Initialize $\pi_\phi$ and $Q_\psi$ with parameters $\phi$ and $\psi$\\
Initialize $\pi_{\bar{\phi}}$ and $Q_{\bar{\psi}}$ with parameters $\bar{\phi} \leftarrow \phi$ and $\bar{\psi} \leftarrow \psi$ \\
Initialize replay buffers $\mathcal{B}$ and $\mathcal{D}$, and success rate buffer $\mathcal{Z}$\\

\For{$\text{episode} = 1$, $M$}
{
    Initialize a random process $\mathcal{N}$ and obtain $o_1$\\
    \For{$t=1$, $T$}
    {
        Get success rate estimate $z$ from $\mathcal{Z}$ \\
        $a_t=\pi_\phi(o_t) + \mathcal{N}_t$\\
        
        \uIf{BC}
        {   
            $a'_t=\pi_\text{expert}(o'_t)$\\
            Save tuple $(o_t, a'_t)$ in $\mathcal{D}$\\
            Sample $\{(o_i, a_i)\}_{i=1}^N$ from $\mathcal{D}$\\
            Set $J_{\text{IL}}(\pi_\phi) = \frac{1}{N}\sum_{i=1}^N(\pi_\phi(o_i) -a_i)^2$\\
        }
        \ElseIf{COACH}
        {
            Query feedback $h_t$\\
            Save tuple $(o_t, a_t, h_t)$ in $\mathcal{D}$\\
            Sample $\{(o_i, a_i, h_i)\}_{i=1}^N$ from $\mathcal{D}$\\
            Set $J_{\text{IL}}(\pi_\phi) = \frac{1}{N}\sum_{i=1}^N(\pi_\phi(o_i) - \clip\left\{a_i+e\cdot h_i, a_\text{min}, a_\text{max}\right\})^2$\\
        }
        Execute $a_t$ and obtain $r_t$ and $o_{t+1}$\\
        Save transition $(o_t, a_t, r_t, o_{t+1})$ in $\mathcal{B}$\\
        
        Sample $\{(o_j, a_j, r_j, o_{j+1})\}_{j=1}^N$ from $\mathcal{B}$\\
        
        Compute $y_j=\begin{cases}
            r_j & \text{if } s_{j+1} \text{ is terminal,} \\
            r_j + \gamma Q_{\bar{\psi}}(o_{j+1}, \pi_{\bar{\phi}}(o_{j+1})) & \text{otherwise.}
        \end{cases}$\\
        Update $Q_\psi$ by minimizing $\frac{1}{N}\sum_{j=1}^N(y_j - Q_\psi(o_j, a_j))^2$\\

        \If{Adaptive $\lambda$}
        {
            Update $\lambda$ by minimizing $J_\lambda(\lambda)$ \\
            $\lambda \leftarrow \max\{\lambda, \lambda_\text{min}\}$
        }
        Update $\pi_\phi$ by minimizing $J_\text{PModL}(\pi_\phi)$: $-z\frac{1}{N}\sum_{j=1}^NQ_\psi(o_j, \pi_\phi(o_j)) + \lambda(1-z)J_{\text{IL}}(\pi_\phi)$\\
        
        Update target networks $Q_{\bar{\psi}}$ and $\pi_{\bar{\phi}}$: $\bar{\phi}\leftarrow \tau \phi + (1-\tau) \bar{\phi}$; 
        $\bar{\psi}\leftarrow \tau \psi + (1-\tau) \bar{\psi}$
        
    }
    Save $\text{success(\textit{episode})}$ in $\mathcal{Z}$
}
\caption{DDPG+PModL}
\label{algo:ddpg+pmodl}
\end{algorithm}

It is worth noting that DDPG+PModL is not explicitly linked to the local planning problem; by considering an appropriate expert and a suitable definition for $\text{success}(episode)$\footnote{Given the episodic settings described in Section~\ref{subsec:episodic_settings}, for the local planning problem (as formulated in this work), $\text{success}(episode)$ is defined by $\mathds{1}_{\left\{\rho^t_\text{target} < \rho_\text{thresh}\right\}} \in \{0, 1\}$.}, the algorithm is readily applicable to different problems.

Moreover, it is important to remark that, if the COACH objective is utilized, the corrective advice that the expert provides to the agent is no longer valid after updating the policy. In practice, as empirically shown by~\cite{perez2020interactive}, we can use a small replay buffer $\mathcal{D}$ to store the expert feedback. Furthermore, in the proposed implementation, the labels for the COACH loss are bounded by the minimum and maximum values of the action components, that is, for a given action $a_i$ and feedback $h_i$, the corresponding label is set to $\clip\left\{a_i+e\cdot h_i, a_{\text{min}}, a_{\text{max}}\right\}$.

\section{Experimental results for the local planning problem}
\label{sec:experimental_results}
\subsection{Experimental configuration}

\begin{figure}
    \centering
    \includegraphics[width=0.8\linewidth]{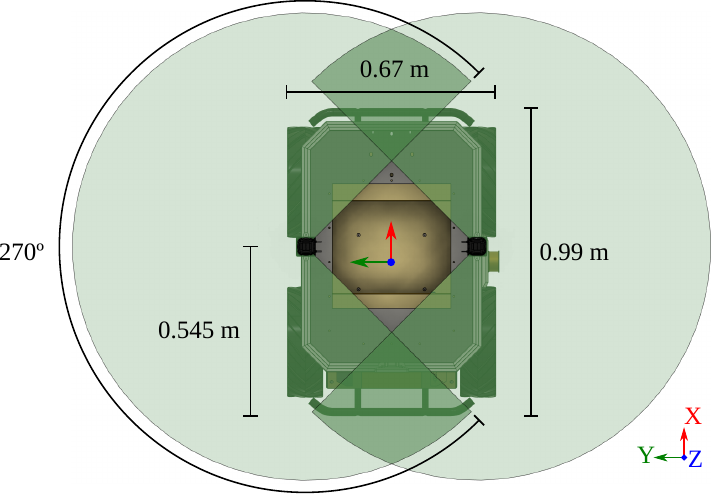}
    \caption{Top view illustration of the Husky A200\textsuperscript{TM}, its dimensions, local frame, and the field of view of the 2D LiDARs installed on the robot.}
    \label{fig:husky_diagram}
\end{figure}

The training process is conducted in simulations, using Gazebo \citep{koenig2004design} and ROS Noetic \citep{quigley2009ros}. A validation of the obtained policies is performed both in simulations and in the real world. The platform chosen to perform the experiments is the Husky A200\textsuperscript{TM} skid-steered mobile robot, equipped with two Hokuyo UTM-30LX-EW 2D LiDARs. Figure~\ref{fig:husky_diagram} shows a top view diagram of the robot and the LiDARs installed on it. 

The $o_\text{pcl}$ set is constructed by transforming the range measurements of each LiDAR to a point cloud referenced to the local frame of the robot, merging these point clouds together, and subsampling the resulting point cloud using a voxel grid filter of size 0.7 m. During training, a ground-truth localization system is utilized to compute $o_\text{target}$, whilst for real-world deployment, the AMCL ROS package is utilized. For the generation of collision free paths, $\mathcal{C}_t$ (see Section~\ref{subsec:synthesizing_il_lp}), Dijkstra's algorithm \citep{dijkstra2022note} is employed.

\subsection{Addressing the simulation to reality gap}
\label{subsec:reality_gap}

Since the platform for deployment is a skid-steered robot, a challenge to reduce the simulation to reality gap is to model its skidding accurately. This is an issue that has to be addressed prior to training so as to facilitate a zero-shot deployment of the learned policies. Furthermore, this task is difficult since skid-based rotations depend on the characteristics of the terrain in which the robot navigates. To account for the above, we propose (i) measuring the real robot's skidding, and (ii) using the acquired data to compute an extended footprint for its differential, simulated counterpart. The extended footprint of this differential robot, after responding to a velocity command, would likely contain the real robot's collision geometry after responding to the same command as an skid-steered robot.

\subsubsection{Measuring skidding}
To measure skidding, angular velocity commands of varying magnitude are sent to the robot when it is over different surfaces, and its pose is estimated using an ArUco marker detector~\citep{garrido2016generation} aligned with its rotation centre. We consider two different surfaces (smooth cement floor and rough floor), and eight different angular speed commands: $\pm$0.25, $\pm$0.5, $\pm$0.75, and $\pm$1.0 rad/s. Each command is sent to the robot until it performs four rotations, and its estimated poses are recorded.

\subsubsection{Computing the extended footprint}
\begin{figure}
    \centering
    \includegraphics[width=\linewidth]{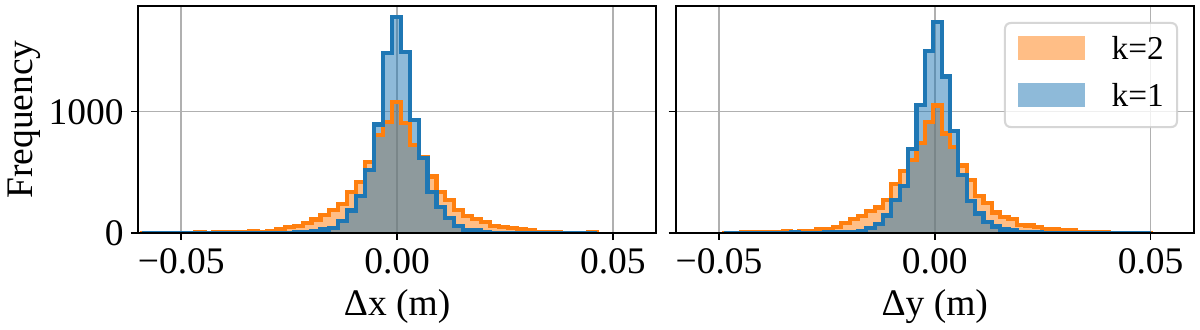}
    \caption{Displacement histograms obtained for the positions estimated at $t+k$ and $t$, taking the pose at $t+k$ as the reference coordinate system.}
    \label{fig:displacement_histogram}
\end{figure}

Due to skidding, the robot changes both its position and orientation when responding to pure angular speed commands, contrary to an ideal differential robot, which would only change its orientation. Given the above, we measure the difference between the robot's position in $t+k$ and in $t$, using the pose in $t+k$ as the reference coordinate system. The histograms of displacements for $k=\{1,2\}$ in the $x$ and $y$ directions, $\Delta x$ and $\Delta y$, are shown in Fig.~\ref{fig:displacement_histogram}. While $k=1$ corresponds to the displacements observed after a single pose detection (at $\sim$11 Hz), due to the robot's control frequency used during training (5 Hz, see Sections~\ref{subsec:training_sim} and~\ref{subsec:validation_sim}), we use the displacements for $k=2$ to select the footprint for the differential robot. Thus, given the obtained measurements, the inflated footprint should be approximately $0.1$ m larger than the real footprint in the $x$ and $y$ directions. In practice, the simulated differential robot's footprint is set to be $1.1 \times 0.8$~m, and maintains the original robot's local frame and the pose of its LiDARs. 

\begin{figure}
    \centering
    \subfloat[Training environment.]{\includegraphics[width=0.45\linewidth]{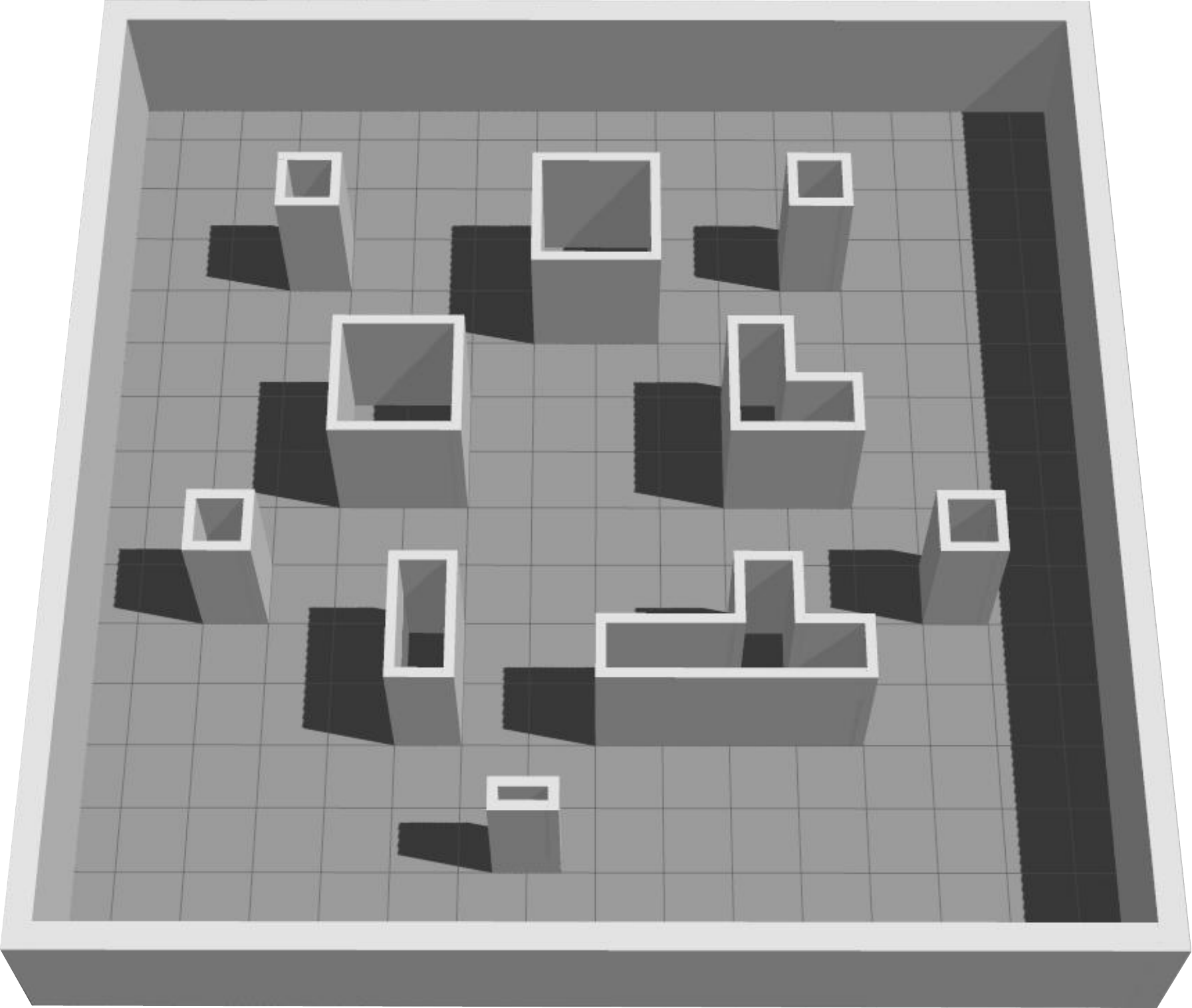}\label{fig:training_env_sim}} \quad
    \subfloat[Validation environment.]{\includegraphics[width=0.45\linewidth]{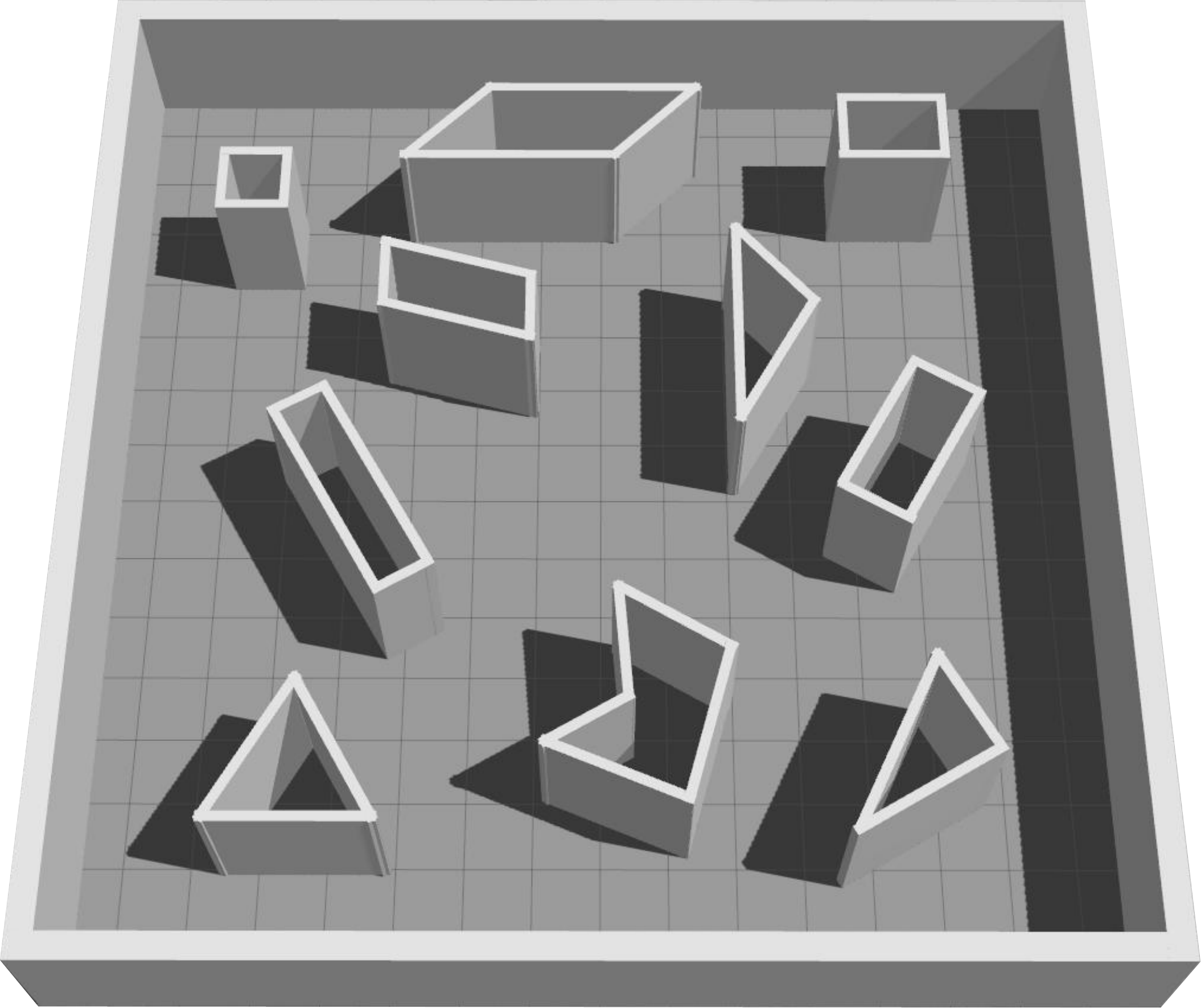}\label{fig:validation_env_sim}} 
    \caption{Environments utilized to train, evaluate and validate the performance of agents in simulations. Both environments are 16 m long by 16 m wide.}
    \label{fig:simulated_environments}
\end{figure}

\subsection{Training in simulations}
\label{subsec:training_sim}

The training process is conducted in the environment shown in Fig.~\ref{fig:training_env_sim}, using a differential robot with extended footprint (see Section~\ref{subsec:reality_gap}). This environment is similar to that proposed in the work of~\cite{xie2018learning} and utilized by~\cite{leiva2020robust}, but scaled from 8$\times$8~m to 16$\times$16~m due to the differences in size between the robots used by~\cite{xie2018learning} and by~\cite{leiva2020robust}, and the Husky A200\textsuperscript{TM}.

We train agents using DDPG, COACH, DAgger, and the proposed method's variants, DDPG+PModL$_{\text{BC}}$ and DDPG+PModL$_{\text{COACH}}$ (see Algorithm~\ref{algo:ddpg+pmodl}). The parameters for these algorithms, the pure pursuit path tracker, and those utilized for the problem formulation, are presented in Table~\ref{tab:system_parameters}. 

\begin{table}
    \centering
    \small
    \caption{Algorithms and problem formulation parameters.}
    \label{tab:system_parameters}
    \resizebox{\linewidth}{!}{%
    \begin{tabular}{llc}
    \toprule
        & \textbf{Parameter} & \textbf{Value}  \\
        \midrule
        \multirow{7}{*}{DDPG} & Actor \& critic learning rates  &  0.0001, 0.001\\
             & Critic weight decay                               &  0.01 \\
             & Smoothing factor $\tau$                           &  0.001 \\
             & Discount factor $\gamma$                          &  0.99 \\
             & Replay buffer size                                &  400,000\\
             & Batch size                                        &  256\\
             & OU noise ($\mu,\sigma, \theta, \Delta t$)         &  0.0, 0.3, 0.15, 0.2  \\
        \midrule
        \multirow{3}{*}{DAgger}      & Policy learning rate      & 0.0001 \\
             & Replay buffer size                                &  - \\
             & Batch size                                        &  256\\
            
        \midrule
        \multirow{5}{*}{COACH}     & Policy learning rate        & 0.0001\\
             & Replay buffer size                                &  256 \\
             & Batch size                                        &  256 \\
             & Error factor $e$ &  0.5\\
             & $\Delta v^{\text{thresh}}_{\text{cmd}-x}$, $\Delta v^{\text{thresh}}_{\text{cmd}-\theta}$ & 0.1, 0.1\\
        \midrule
        \multirow{3}{*}{PModL} & $K_{\text{SR}}$ & 100\\
             & $\lambda$ learning rate & 0.025\\
             & $\lambda_\text{min}$ & 1\\
        \midrule
        Pure Pursuit & $\rho_\text{look-ahead}$, $K_x$, $\theta^{+}_\text{wp}$, $\theta^{-}_\text{wp}$ & 0.5, 2, $\frac{\pi}{6}$, $\frac{\pi}{2}$ \\
        \midrule
        Observations & $\rho_\text{max}, \rho^{\text{max}}_\text{target}$ & 18.0, 18.0 \\
        \midrule
        \multirow{3}{*}{Control}  & Control frequency (Hz) & 5 \\
             & $v^{\text{min}}_{\text{cmd}-x}$, $v^{\text{max}}_{\text{cmd}-x}$ (m/s) & $\; \; \;$0, 1\\
             & $v^{\text{min}}_{\text{cmd}-\theta}$, $v^{\text{max}}_{\text{cmd}-\theta}$ (rad/s)& $-$1, 1\\
        \midrule
        \multirow{4}{*}{\begin{tabular}[c]{@{}l@{}}Reward\\ function\end{tabular}} & $r_\text{success}$, $r_\text{collision}$, $\theta_\text{fov}$ & 100, $-$100, $\frac{2}{3}\pi$ \\
             & $K_\text{speed}$, $K_\text{target}$  & 1.0, 5.0\\
             & $K'_\text{target}$, $K_\text{fov}$, $K'_\text{fov}$ & 6.0, 3.0, 5.0\\ 
             & $K_\text{danger}$, $\rho_\text{danger-thresh}$ & $-$10.0, 0.7\\
        \midrule 
        \multirow{2}{*}{\begin{tabular}[c]{@{}l@{}}Episodic\\ settings\end{tabular}} & $\rho_\text{init-thresh}$ & 3.0\\
             &  $\rho_\text{thresh}$ & 0.2\\
    \bottomrule
    \end{tabular}
    }
\end{table}

\begin{figure*}
    \centering
    \includegraphics[width=\linewidth]{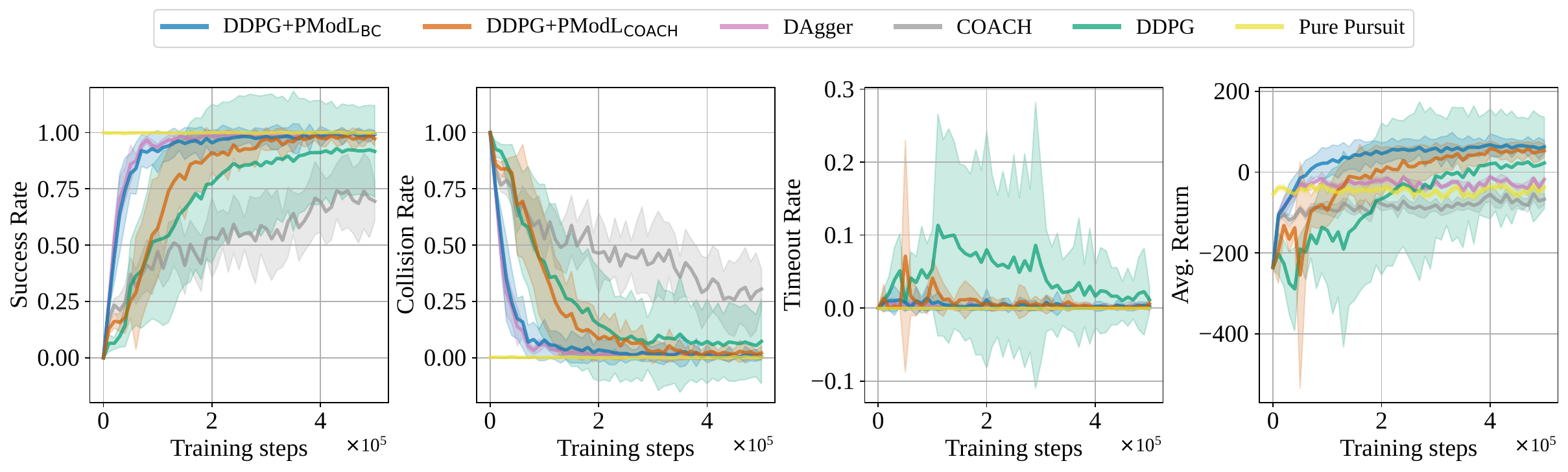}
    \caption{Performance evolution of the agents during the training process. All agents are evaluated every 1$\times$10$^4$ training steps for 100 episodes. For agents trained using DDPG and the DDPG+PModL variants, the exploration noise is turned off to conduct this evaluation. Furthermore, the random seed controlling the episodic settings is fixed per trial. For each reported metric, the solid curves correspond to its average value across eight training trials, whilst the shaded area corresponds to its standard deviation.}
    \label{fig:training_perf_evolution}
\end{figure*}

\begin{figure*}
    \centering
    \includegraphics[width=\textwidth]{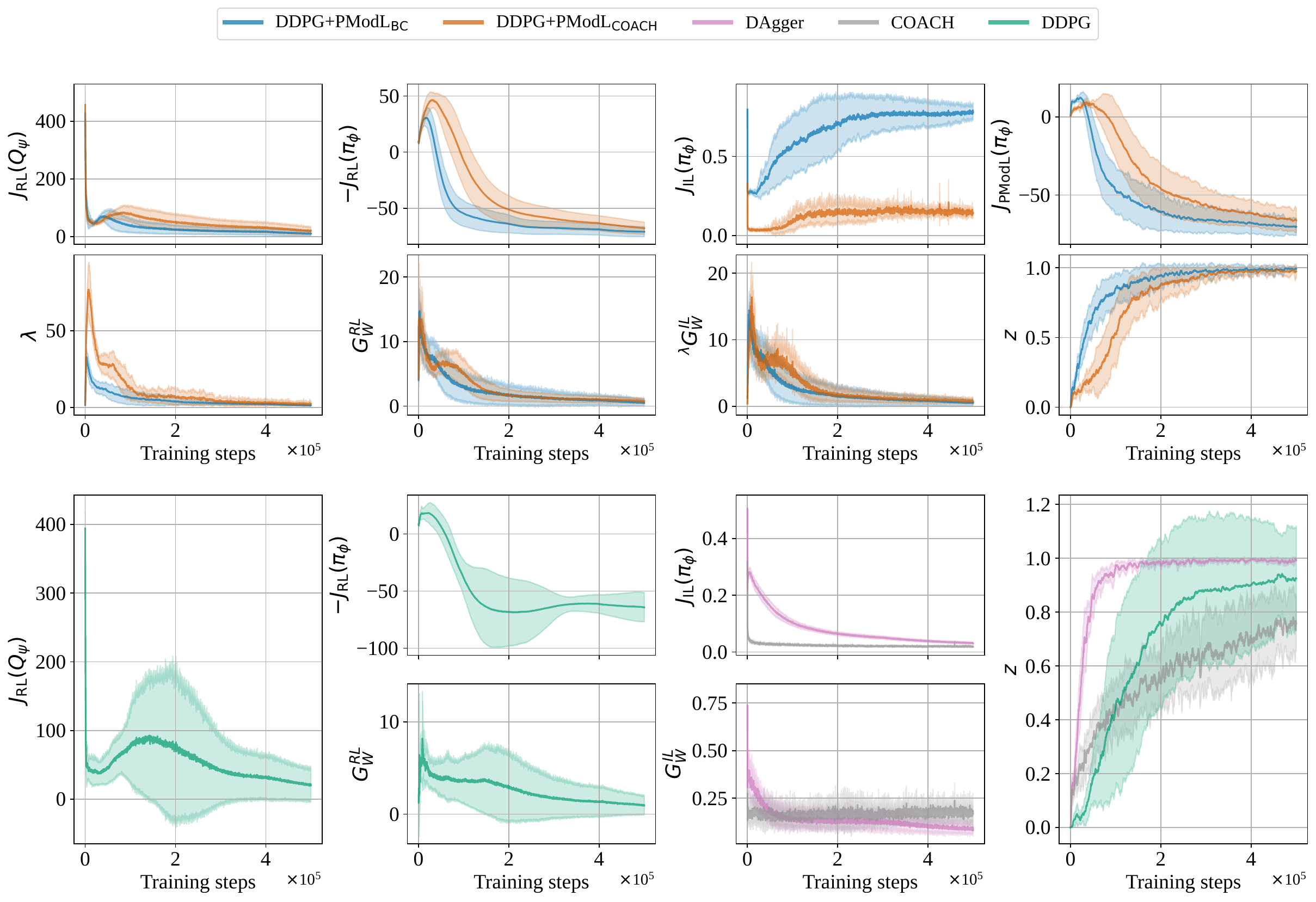}
    \caption{Evolution of the value of loss functions, norm of gradients and success rate estimates $z$ during training, for all agents. For the DDPG+PModL variants, the evolution of $\lambda$ is also included. For each training trial, the reported metrics are averaged and logged every 100 training steps. The solid curves correspond to the average value of these logs across eight training trials, and the shaded areas to their standard deviation.}
    \label{fig:training_losses_evolution}
\end{figure*}

\begin{figure*}[h]
    \centering
    \includegraphics[width=\textwidth]{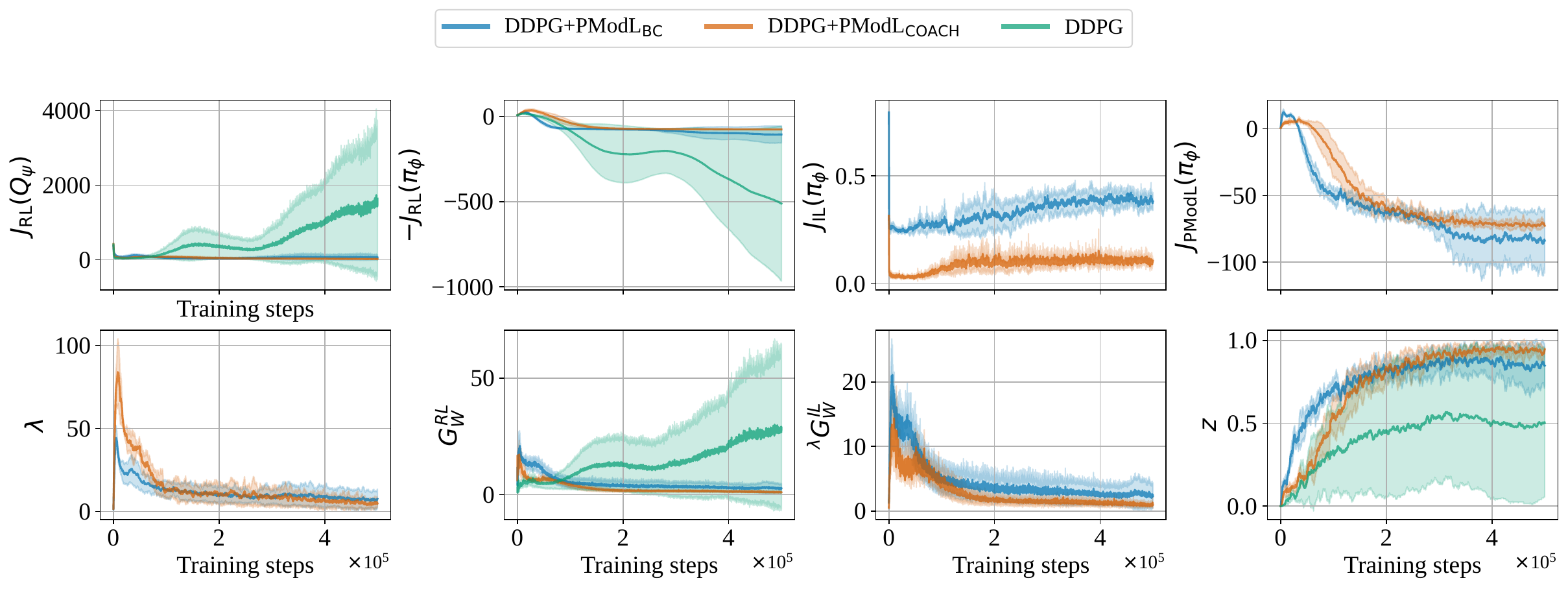} 
    \caption{Evolution of the value of loss functions, norm of gradients, success rate estimates $z$, and $\lambda$ factors during training for DDPG, and the DDPG+PModL variants. The curves are constructed as described in Fig.~\ref{fig:training_losses_evolution}. In this case the training process is conducted using a reward function similar to that described in Section~\ref{subsec:reward_function}, but that does not have the $r^t_{\text{fov}}$ nor the $r^t_{\text{danger}}$ terms (Eqs.~\eqref{eq:r_fov} and~\eqref{eq:r_danger}), making it less informative.}
    \label{fig:ablation_reward}
\end{figure*}
 
DDPG+PModL$_{\text{BC}}$ inherits the parameters of DDPG and DAgger, whilst DDPG+PModL$_{\text{COACH}}$ inherits the parameters of DDPG and COACH. The training process is conducted for 5$\times$10$^5$ steps for all agents. For DDPG and the DDPG+PModL variants, the Ornstein-Uhlenbeck (OU) noise is modulated by a factor that decays linearly from 1.0 to 0.05 in 4$\times$10$^5$ steps. For each algorithm, eight independent training trials are performed, and in all cases the number of training steps is equal to the number of agent-environment interactions that are simulated.

The evolution of the agents' performance in their training environment, measured in terms of success rate, collision rate, timeout rate (400 steps maximum per episode), and average return, is shown in Fig.~\ref{fig:training_perf_evolution}. Additionally, the evolution of losses, norms of gradients (see Section~\ref{subsubsec:computing_lambda}), success rate estimates $z$, and balancing factors $\lambda$ (for the DDPG+PModL variants) are shown in Fig.~\ref{fig:training_losses_evolution}.  

From the performance evolution curves, it is observed that in terms of pure success rate, the agents trained using DAgger and the DDPG+PModL variants outperform the agents trained using DDPG and COACH. The dispersion on all the obtained metrics is also lesser for DAgger and the DDPG+PModL variants compared to COACH and DDPG, the latter being the algorithm presenting the highest performance variability across different training trials. The average return curves, on the other hand, show that only the RL-based approaches surpass zero return. This result is expected, since agents trained with said approaches aim to maximize this metric, while the COACH and DAgger agents are just receiving learning signals derived from the pure pursuit path tracker. It is worth noting that DAgger agents achieving higher average returns than those controlled using pure pursuit by the end of training, simply indicates that their learned behavior (not being identical to that established by pure pursuit), by chance, is better suited to maximize the reward function defined in Section~\ref{subsec:reward_function}.

From Fig.~\ref{fig:training_perf_evolution}, it is also observed that DDPG+PModL$_{\text{BC}}$ and DAgger have the best success rate evolutions (fast convergence to a perfect score), and both algorithms present the lowest dispersion across trials. Furthermore, although pure COACH does not get to a good success rate by the end of the training process, its combination with RL using the proposed method (DDPG+PModL$_{\text{COACH}}$) almost matches the success rate of DAgger and DDPG+PModL$_{\text{BC}}$ at 5$\times$10$^5$ training steps.

We can also observe that pure RL does not get to match the performance, in terms of average success rate, of the DDPG+PModL agents; in fact, the DDPG+PModL$_\text{BC}$ agents achieve the average success rate of the DDPG agents using approximately 4 times less experiences.

Fig.~\ref{fig:training_losses_evolution} reveals important information regarding the mechanisms incorporated in the proposed method to balance and modulate losses over time, and also shows how losses and norms of gradients evolve for pure RL (DDPG) and pure IL/IIL (DAgger/COACH). 

The effect of a varying $\lambda$ factor that balances the actor's RL and IL losses is evidenced by looking at the norms of the backpropagated gradients, $G^{\text{RL}}_{W}$ and $^{\lambda}G^{\text{IL}}_{W}$, for the DDPG+PModL variants. As it can be observed, the evolution of these norms is such that they are constantly numerically similar as the training process progresses. In contrast, the numerical values of $^{\lambda}G^{\text{IL}}_{W}$, for the DDPG+PModL variants, greatly differs from those of $G^{\text{IL}}_{W}$, for DAgger and COACH. The effect of discrepancies between these norms is further discussed in Appendix~\ref{appendix:lambda_ablation}, where the proposed automatic adjustment for $\lambda$ is compared to setting it to default values throughout the whole training process.

Fig.~\ref{fig:training_losses_evolution} also reaffirms that a faster and more stable convergence is achieved using either DDPG+PModL variant compared to pure RL. The dispersion of $z$ across different training trials observed for the DDPG agents suggests that some of them present a success rate evolution similar to those obtained by DDPG+PModL$_\text{BC}$ agents. The above, however, can be attributed to the reward function utilized for training, which has been hand crafted and fine tuned carefully, taking into account a fair amount of expert knowledge. In the following, we show the huge effect that has making the reward function less informative, and how the proposed method still achieves almost perfect scores on success rate, whereas pure RL fails to converge consistently.

\subsubsection{Training using a less informative reward function}

Fig.~\ref{fig:ablation_reward} shows the evolution of losses, norm of gradients and success rate estimates for agents trained using DDPG and the DDPG+PModL variants, using a slightly modified reward function, obtained by eliminating the $r^t_{\text{fov}}$ and the $r^t_{\text{danger}}$ terms (Eqs.~\eqref{eq:r_fov} and~\eqref{eq:r_danger}) from the original function defined in Eq.~\eqref{eq:reward_function}. For each algorithm, four independent training trials are conducted.

The described reward function modification, although minor, has a huge impact on the performance attained for agents trained using pure RL (DDPG), which do not get to converge in all of the trials. The agents trained using the DDPG+PModL variants, in contrast, show metrics that are similar to those observed when training using the full reward function (compare Fig.~\ref{fig:ablation_reward} with Fig.~\ref{fig:training_losses_evolution}). Moreover, in this case the DDPG+PModL$_{\text{COACH}}$ agents achieve a better asymptotic performance than those trained using DDPG+PModL$_{\text{BC}}$ (in average), however, this could be attributed to a lesser number of performed training trials.

\begin{table*}[h]
    \small
    \centering
    \caption{Performance of the trained agents when deployed in the simulated validation environment, averaged across eight trials per algorithm.}
    \label{tab:sim_validation}
    \resizebox{\textwidth}{!}{%
    \begin{tabular}{llcccccc}
    \toprule
    \textbf{Robot} & \textbf{Algorithm} & \textbf{SR} & \textbf{CR} & \textbf{TR} & \textbf{SPL} & \textbf{Avg. Return} & \textbf{Avg. Steps} \\
    \midrule
    \multirow{5}{*}{\begin{tabular}[c]{@{}l@{}}Differential\\ (Augmented \\ footprint)\end{tabular}} & DDPG  & 0.8520$\pm$0.2031 & 0.1293$\pm$0.1696  & 0.0188$\pm$0.0360 &0.7903$\pm$0.1876  & $-$4.025$\pm$86.43 & 53.14$\pm$22.18 \\
    & DAgger  & 0.8415$\pm$0.0430 & 0.1435$\pm$0.0484  & 0.0150$\pm$0.0083 &0.8362$\pm$0.0423  & $-$41.16$\pm$25.03 & 45.29$\pm$3.994  \\
    &COACH  & 0.6252$\pm$0.1278 & 0.3737$\pm$0.1276  & 0.0010$\pm$0.0017 & 0.6107$\pm$0.1209 & $-$72.05$\pm$16.65  & 38.54$\pm$5.357  \\
    & DDPG+PModL$_{\text{COACH}}$ & 0.8957$\pm$0.0985 & 0.0980$\pm$0.0973  & 0.0063$\pm$0.0047 &0.8353$\pm$0.0902  & $+$16.71$\pm$34.09 & 46.09$\pm$3.728  \\
    & DDPG+PModL$_{\text{BC}}$ & 0.9590$\pm$0.0778 & 0.0378$\pm$0.0725  & 0.0033$\pm$0.0058 &0.8957$\pm$0.0726  & $+$45.63$\pm$37.57 & 43.48$\pm$2.046  \\
    \midrule
    \multirow{2}{*}{\begin{tabular}[c]{@{}l@{}} Husky (SCF)$^{a}$ \end{tabular}} & DDPG+PModL$_{\text{COACH}}$ & 0.8445$\pm$0.0922 & 0.1555$\pm$0.0922 & 0.0000$\pm$0.0000   & 0.7656$\pm$0.0819  & $+$10.54$\pm$28.43 & 43.81$\pm$1.579  \\
    & DDPG+PModL$_{\text{BC}}$ & 0.9187$\pm$0.0703 & 0.0813$\pm$0.0703 & 0.0000$\pm$0.0000   & 0.8364$\pm$0.0650  & $+$34.40$\pm$24.75 & 43.81$\pm$0.646  \\
    \midrule
    \multirow{2}{*}{\begin{tabular}[c]{@{}l@{}} Husky (HCF)$^{b}$\end{tabular}} & DDPG+PModL$_{\text{COACH}}$ & 0.8740$\pm$0.0982 & 0.1258$\pm$0.0982  & 0.0003$\pm$0.0007 & 0.8040$\pm$0.0890  & $-$258.9$\pm$105.3  & 195.7$\pm$12.34 \\
    & DDPG+PModL$_{\text{BC}}$ & 0.9455$\pm$0.0702 & 0.0545$\pm$0.0702  & 0.0000$\pm$0.0000 & 0.8736$\pm$0.0662  & $-$172.2$\pm$99.39  & 189.3$\pm$2.134 \\
    \bottomrule
    \multicolumn{8}{l}{\begin{tabular}[c]{@{}l@{}l} \footnotesize{$^{a}$} & \footnotesize{Standard control frequency (5 Hz).} \\
    \footnotesize{$^{b}$} & \footnotesize{High control frequency (50 Hz).}
    \end{tabular}}
    \end{tabular}
    }
\end{table*}

\begin{figure*}[h!]
    \centering
    \includegraphics[width=0.8\textwidth]{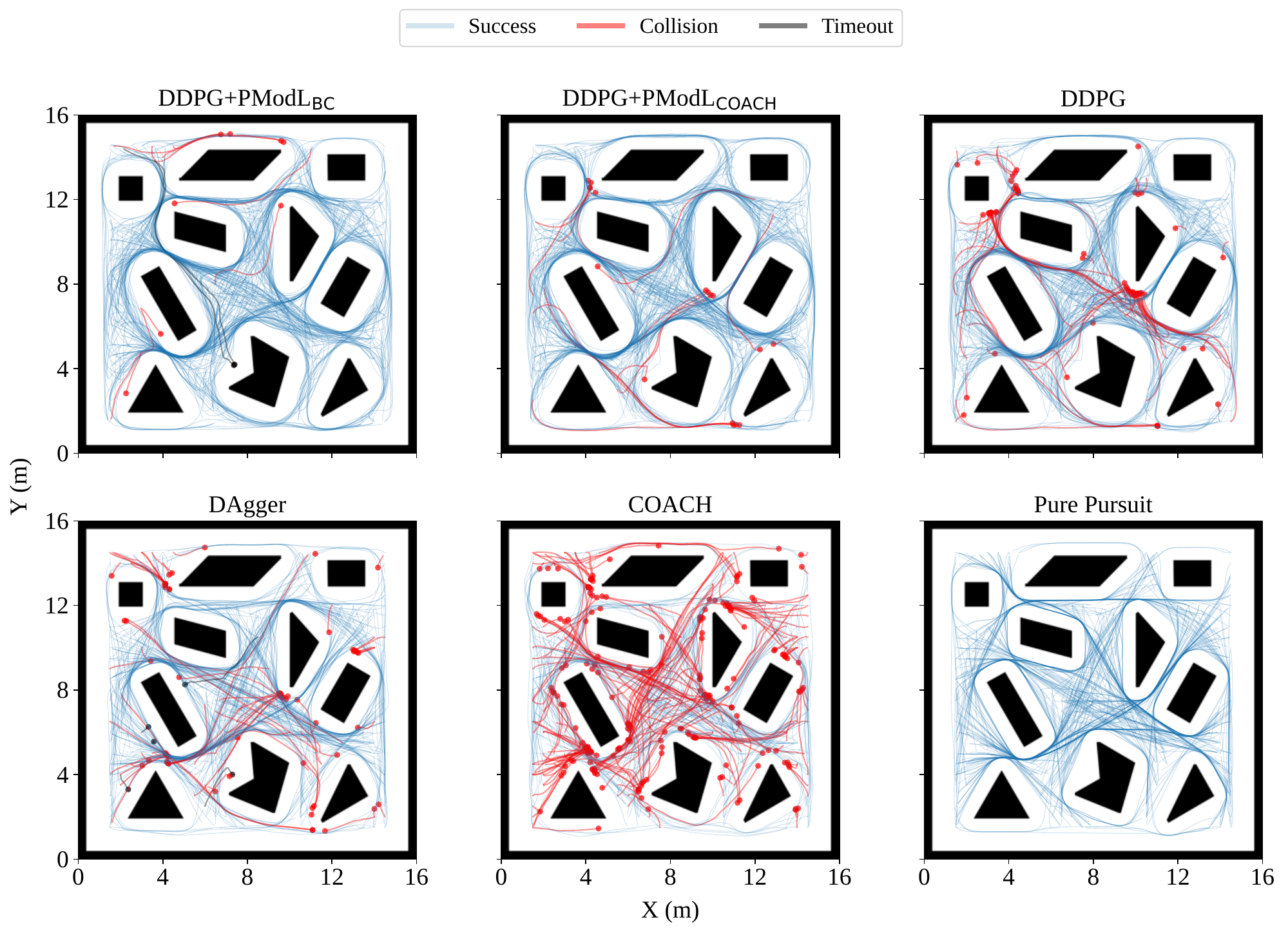}
    \caption{Trajectories followed by the trained agents, and an agent controlled directly by the pure pursuit path tracker, when deployed in the simulated validation environment. The trajectories displayed correspond to those executed in 500 episodes by a single agent (that is, by the policy obtained after a single training trial). The blue trajectories correspond to successful episodes, the red ones to those where a collision occurred, and the black ones to those that ended in a timeout.}
    \label{fig:sim_val_trajectories}
\end{figure*}

The above can be explained considering that the exploration problem difficulty increases without $r^t_{\text{danger}}$ and $r^t_{\text{fov}}$ as part of the reward function definition, as these terms encourage the agent to stay away from obstacles, and to avoid orientations that would increase its distance with respect to the navigation goal if driving forward, respectively. The guidance provided by these terms, specially by $r^t_{\text{danger}}$ (since $r_{\text{speed}}$ also takes into account the agent's orientation with respect to the goal), seem to be crucial to consistently learn acceptable policies using DDPG.

The near perfect success rate achieved by the agents trained using DDPG+PModL can be explained considering that the exploration challenge in this case is alleviated by the availability of expert guidance, via demonstrations or feedback, depending on the utilized algorithm's variant.

\subsection{Validation in simulations}
\label{subsec:validation_sim}

To validate the performance of these agents in simulations, they are deployed in the environment shown in Fig.~\ref{fig:validation_env_sim}, and each of them is evaluated for 500 episodes, with a fixed random seed controlling the episodic settings. The validation is conducted using the same ideal differential robot used for training (which has an augmented footprint), and using a simulated Husky A200\textsuperscript{TM} with the standard 5 Hz control frequency (SCF) used for training, and with a higher control frequency (HCF), set to 50 Hz. The quantitative results for these experiments are shown in Table~\ref{tab:sim_validation}, whereas some of the trajectories performed by the agents are shown in Fig.~\ref{fig:sim_val_trajectories}. For Table~\ref{tab:sim_validation}, SR stands for success rate, CR for collision rate, TR for timeout rate, and SPL for success weighted by path length \citep{anderson2018evaluation}. The latter is defined as $\frac{1}{M_{\text{eval}}}\sum^{M_{\text{eval}}}_{i=1}\mathds{1}_{\{S_i\}}\frac{l^{\text{shortest}}_i}{\max\{l^{\text{shortest}}_i, l^{\text{agent}}_i\}}$, where $M_\text{eval}$ is the number of evaluation episodes, and for each episode $i$, $S_i$ indicates whether or not said episode was successful, $l^{\text{shortest}}_i$ is the shortest path length from the agent's starting position to the navigation target, and $l^{\text{agent}}_i$ is the actual path length traversed for the agent in said episode.

For this new environment, the obtained metrics show a performance detriment for all of the trained agents. This is expected, given that the validation is performed in an environment that produces out-of-distribution observations. The decrease in performance of agents trained using RL methods (DDPG and the DDPG+PModL variants), however, is minimal compared with the rest of the agents; for instance, the success rate of agents trained using DAgger and COACH decreases greatly compared to the rate measured in the training environment. 

The above showcases that the performance that local planning agents get in their training environment is not necessarily a good proxy for predicting their performance in a different environment. This is specially relevant for the problem addressed in this work, where agents are expected to generalize. 

The metrics obtained also show that agents trained using the DDPG+PModL variants outperform all the others in the success and collision rate metrics (which, as stated previously, was not evident by examining the evaluation metrics on the training environment, see Fig.~\ref{fig:training_perf_evolution}); the best performance overall being attained by the agents trained using DDPG+PModL$_\text{BC}$, followed by those trained using DDPG+PModL$_\text{COACH}$. In fact, it is observed that, in terms of average success rate, the DDPG+PModL$_\text{BC}$ agents outperform those trained using pure RL (DDPG) by 12.5\%, and those trained using pure IL (DAgger) by 13.9\%, whilst the DDPG+PModL$_\text{COACH}$ agents outperform the DDPG agents by 5.1\%, and the DAgger agents by 6.4\%. It is also worth mentioning that even though the DDPG+PModL$_\text{COACH}$ agents average a slightly worse SPL metric compared to the DAgger agents, they also present better success and collision rate metrics compared to said agents. 

The observed performance difference between DDPG+PModL$_\text{COACH}$ and DDPG+PModL$_\text{BC}$ agents may be explained given that the quality of the feedback they receive is sub optimal by definition. It is possible that this feedback generates learning signals that are contradictory to those produced by pure RL during training. This issue is not as exacerbated for the DDPG+PModL$_\text{BC}$ agents, which receive unmodified demonstrations provided by the pure pursuit path tracker, and although these demonstrations are sub optimal with respect to the reward function, are coming from a scripted policy designed to complete the local planning task. Furthermore, as the DDPG+PModL$_\text{BC}$ agents get to a higher success rate estimate $z$ (see Section~\ref{subsec:iconcorporating_il_to_rl}) faster than those trained using DDPG+PModL$_\text{COACH}$, they get to learn almost purely by RL for a larger fraction of the whole learning process, which may contribute to the observed differences in performance. 

If contradictory learning signals are generated due to lower quality feedback, then this issue could be alleviated by detecting scenarios in which these signals occur, and extending the objective presented in Eq.~\eqref{eq:pmodil_loss} to modify the relative importance between $J_\text{RL}(\pi_\phi)$ and $J_{\text{IL}}(\pi_\phi)$ for such cases.

The trajectories shown in Fig.~\ref{fig:sim_val_trajectories} provide a qualitative insight on the learned behaviors (an the performance) of the agents. For instance, there is a similar pattern on the successful trajectories traversed by agents trained using DAgger compared to those generated by controlling the differential robot using pure pursuit, which is expected since DAgger agents learn to imitate the path tracker. While this pattern is also observed for trajectories associated with successful episodes for agents trained using COACH, these agents possess a high crash rate, since red trajectories are associated with episodes ending in collision. On the other hand, there is a different pattern that is similar for all the agents that are trained using RL-based approaches (DDPG and the DDPG+PModL variants). The main difference between these patterns (those followed by agents trained using pure IL/IIL and those trained using RL) is that there is a noticeable tendency for RL-based approaches to stay as far as possible from the obstacles they encounter (likely because of the $r^t_\text{danger}$ term in the reward function definition), whereas agents trained only relying on the pure pursuit path tracker tend to favor straight and short paths (which for the tracker are generated by a global planner with access to a map of the environment), although that may imply passing closer to obstacles.

Finally, the experiments conducted using the simulated Husky A200\textsuperscript{TM} and the DDPG+PModL variants, show that there is a noticeable overall performance detriment when running the policies at the standard 5 Hz control frequency, and this detriment is substantially alleviated when running the policies at 50 Hz (cf. ``Husky (SCF)'' with ``Husky (HCF)'' in Table~\ref{tab:sim_validation}). 

The higher crash rate observed in these experiments is likely due to skidding, however, as stated before, this effect is diminished by running the controller at a higher frequency. Although the skidding observed in the simulator is greater than in the real-world, empirically a higher control frequency allows the agent to reactively correct these disturbances. The reduction in timeout rates may also be attributed to skidding, as timeouts often occur when agents face situations that trigger a sequence of actions that make the robot oscillate in place. The simulated Husky, however, drifts when commanded alternating angular velocities, thus, eventually colliding with nearby obstacles.

It is worth noting that for the ``Husky (HCF)'' experiments, the maximum number of steps per episode (until a timeout) was increased (to account for a higher control frequency), from 400 to 4000 steps. This change explains the huge discrepancy in the average number of steps and average returns reported in Table~\ref{tab:sim_validation} compared to all the other experiments discussed in this section.

\begin{figure*}[h!]
    \centering
    \subfloat[]{\includegraphics[width=0.22\linewidth]{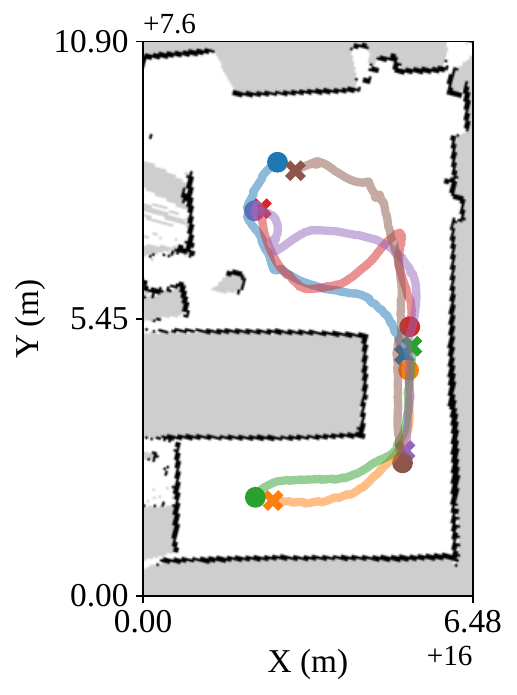} \label{fig:indoors_1}
    \qquad \includegraphics[width=0.7\linewidth]{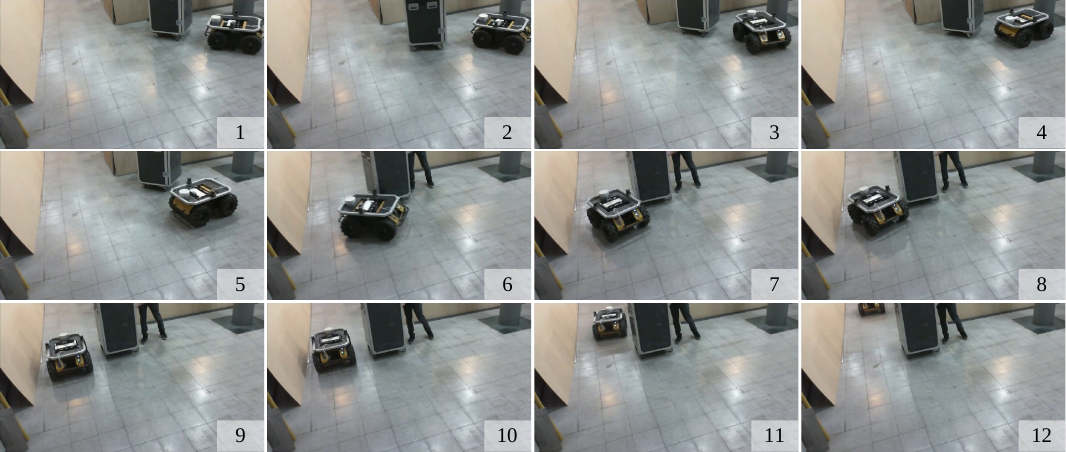}}
    \\
    \subfloat[]{ \includegraphics[width=0.22\linewidth]{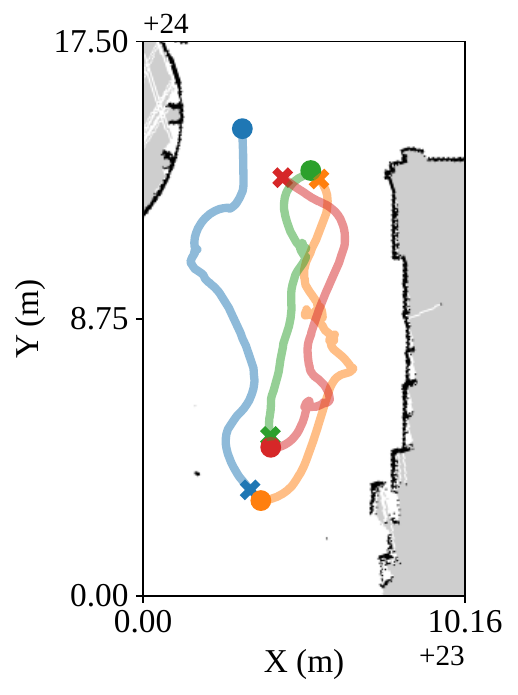}\label{fig:outdoors_2}
    \qquad \includegraphics[width=0.7\linewidth]{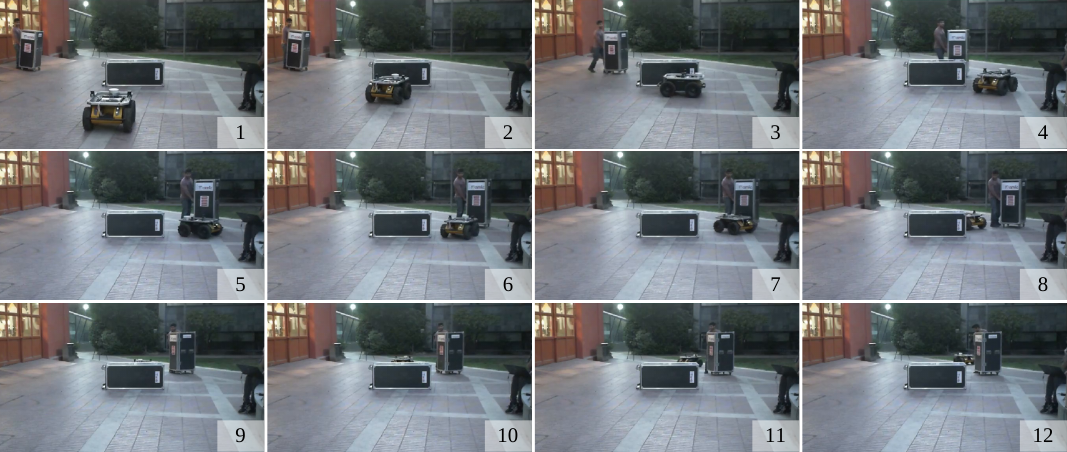} }
    \caption{Examples of trajectories executed by a DDPG+PModL$_\text{BC}$ agent when deployed in (a) an indoor environment, and (b) an outdoor environment. For the trajectories shown at the left of (a) and (b), dots are starting points, whereas crosses correspond to targets. The sequence of images at the right of (a) and (b) correspond to video frames of a single trajectory associated with the maps shown at the left. These frames are sequentially enumerated, and taken at 0.5~Hz.}
    \label{fig:real_world_experiments}
\end{figure*}

\subsection{Real world deployment}

To validate the proposed method in the real world, an agent trained using DDPG+PModL$_\text{BC}$ was used to control a real Husky A200\textsuperscript{TM} robot.\footnote{The agent utilized for these experiments was trained using $\lambda=40$, given that the technique to dynamically adjust $\lambda$ was developed in parallel to real-world testing. As shown in Appendix \ref{appendix:lambda_ablation}, no appreciable differences exist between the average value of $z$ for agents trained using dynamic $\lambda$ and $\lambda =40$.} To reduce the risk of colliding due to skidding, and given the results obtained when validating policies using the simulated Husky robot (see Table~\ref{tab:sim_validation}), a control frequency of 50 Hz was set. Furthermore, a denser point cloud to construct $o_\text{pcl}$ was used to account for a wider variety of obstacles. 

We conducted navigation experiments in simple indoor and outdoor environments, where several navigation targets were queried to the agent sequentially. For some experiments we purposefully intercepted the agent using movable obstacles to evaluate its reactiveness. Some of the trajectories executed by the robot in these environments are shown in Fig.~\ref{fig:real_world_experiments}, whilst video recordings of some of these experiments can be found in~\url{https://youtu.be/mZlaXn9WGzw}. Finally, a quantitative measure of the agent's performance when deployed in the real world is shown in Table~\ref{tab:real_world_exps_results}.

Fig.~\ref{fig:real_world_experiments} shows both several trajectories in the environments they were executed, and frames of a single trajectory in said environments (left and right of Figs.~\ref{fig:indoors_1} and~\ref{fig:outdoors_2}, respectively). Moreover, these frames correspond to experiments in which we purposefully utilized a movable object to check if the agent was capable to avoid it.

The obtained results show that the agent behaves adequately in the real-world, being able to navigate on different terrains, passing through narrow passages and reacting properly to dynamic obstacles. Furthermore, it was able to compensate for localization errors, which were present during some of the conducted experiments, and that resulted in the agent executing sub-optimal trajectories.

\begin{table}[h]
    \small
    \centering
    \caption{Real world performance of the DDPG+PModL$_{\text{BC}}$ agent.}
    \label{tab:real_world_exps_results}
    \resizebox{\linewidth}{!}{%
    \begin{tabular}{ccccc}
        \toprule
        \textbf{Env.} & \# \textbf{Trials} & \textbf{SR} & \textbf{CR/TR} & \textbf{SPL} \\
        \midrule
        Indoor & 21 & 1.00$\pm$0.00 & 0.00$\pm$0.00 & 0.8726$\pm$0.1287 \\
        Outdoor & 8 & 1.00$\pm$0.00 & 0.00$\pm$0.00 & 0.8053$\pm$0.1131\\
        \bottomrule
    \end{tabular}
    }
\end{table}

\section{Conclusion}
\label{sec:conclusion}

In this work, we presented a general method to combine reinforcement learning and imitation learning by means of a dynamic modulation dependent on backpropagated gradients and the estimated performance of the agents during training. We also proposed a practical implementation of this method using an off-policy RL algorithm (DDPG) in combination with IL (BC) or IIL (COACH), and evaluated its application to the local planning problem. For said application, we showed that the proposed method not only accelerates the learning process when compared to pure RL (requiring approximately 4 times less experiences to reach the same success rate during training), but also allows obtaining better policies in a consistent manner, obtaining an average success rate of 0.959 and outperforming pure RL by 12.5\%, and pure IL by 13.9\%. Moreover, we empirically showed that the trained controllers are able to perform effectively in the real world by addressing the sim2real transfer problem using a simple methodology, targeted to allow zero-shot deployment of the obtained policies in skid-steered robots.

It should be noted that the proposed method is not constrained to the specific implementation developed in this work. For instance, a task that could be readily addressed using DDPG+PModL (with a suitable expert) would be the ``reaching'' task (in robotic manipulation). A loose analogy between local planning and reaching can be established, since for reaching the aim is to get the end effector of a manipulator to a certain target pose, while avoiding collisions with the environment and with other links of the manipulator itself. Moreover, although in this work the IIL signals are synthesized online, they may also be provided online by a human. This comes with challenges that are not addressed in this work, such as temporal delay in the feedback given to the agent, however, it relaxes the requirements of the algorithm by no longer needing an expert policy. 

Thus, applying the proposed method to different tasks, using real human feedback, and generalizing it, for instance, to stochastic policies and other existing RL algorithms, is left as future work.

\begin{figure*}[h]
    \centering
    \includegraphics[width=\textwidth]{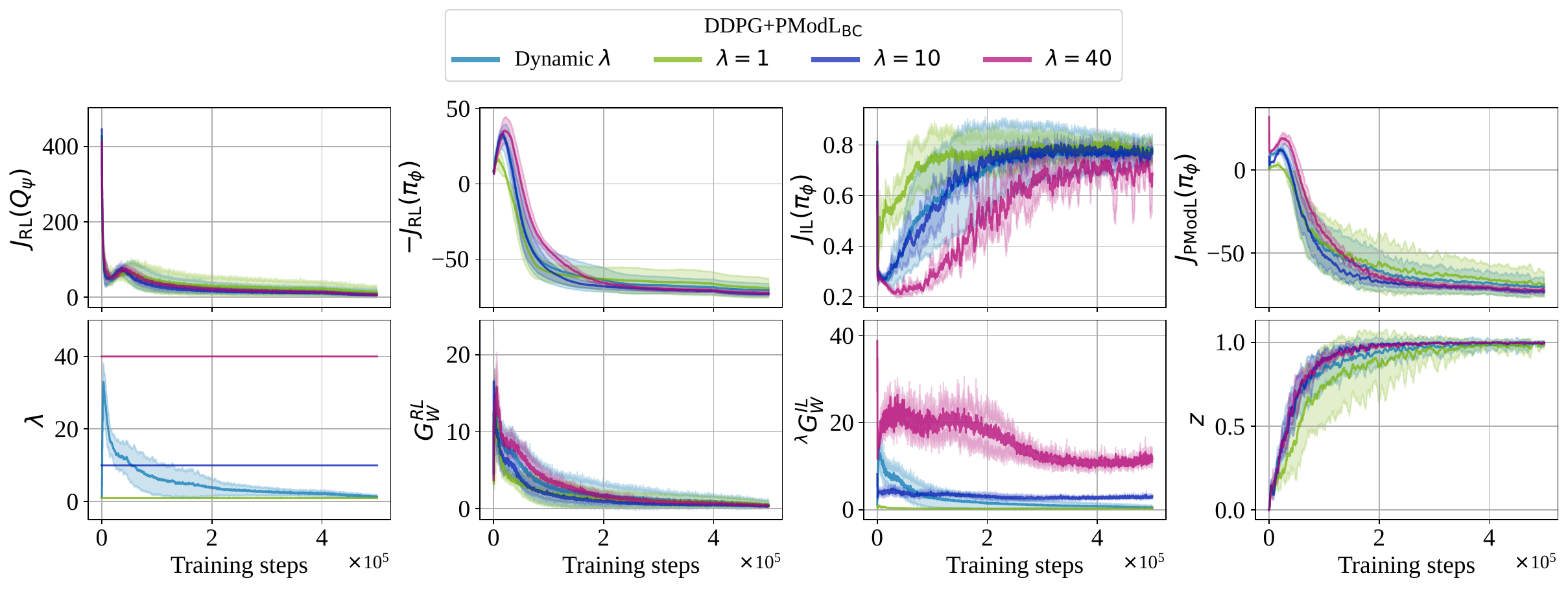} 
    
    \vspace{0.5cm}
    \includegraphics[width=\textwidth]{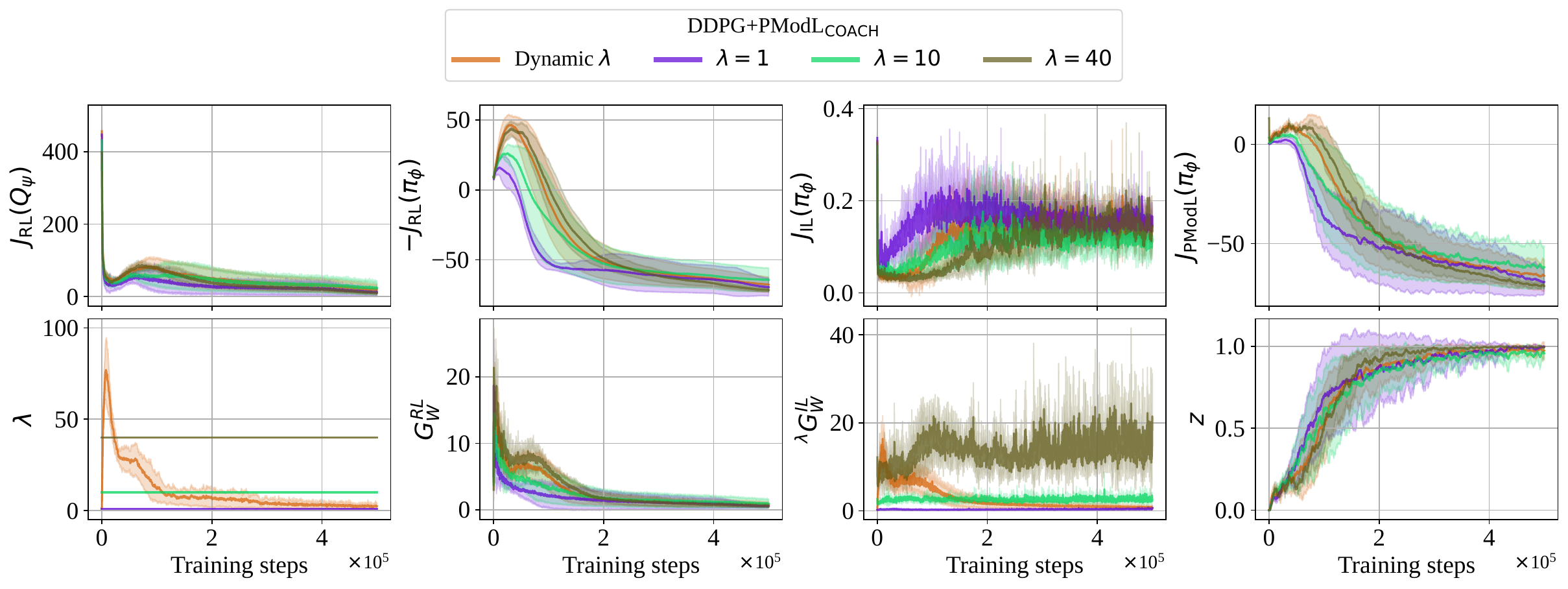}
    \caption{Evolution of the value of loss functions, norm of gradients, success rate estimates $z$, and $\lambda$ factors during training for the DDPG+PModL variants, using dynamic and fixed values for $\lambda$. The curves are constructed as described in Fig.~\ref{fig:training_losses_evolution}.}
    \label{fig:lambda_ablation}
\end{figure*}

\appendix

\section*{The role of $\lambda$ in DDPG+PModL for local planning}
\label{appendix:lambda_ablation}

Figure~\ref{fig:lambda_ablation} shows the evolution of loss functions, norms of gradients, success rate estimates $z$, and $\lambda$ factors for DDPG+PModL variants trained using dynamic and fixed $\lambda$ factors, averaged across four training trials.

The curves show that setting $\lambda=1$ generally results in a higher variability in performance across training trials. 

The differences in the estimated success rate $z$ between setting $\lambda$ to 10, 40 or varying it dynamically for the DDPG+PModL$_{\text{BC}}$ agents is minimal. For the DDPG+PModL$_{\text{COACH}}$ agents, the above also holds true, however, setting $\lambda=10$ also results in a greater dispersion across trials.

These evolution curves show that the proposed method, under the formulation presented in this work, seems to be robust to different schedules for setting $\lambda$ factors. It must be stressed, however, that this may not be the case for different tasks or even for different formulations for the local planning problem due to $G^{\text{RL}}_{\text{W}}$ being linked to the reward function. In this sense, using the proposed automatic adjustment for $\lambda$ would be preferred, as it makes the algorithm agnostic to these design decisions, and does not increase the number of parameters that need to be tuned for the learning algorithm.

\balance

\section*{Acknowledgements}
The authors would like to thank Pablo Alfessi for the technical support he provided to carry out the experimental tests for real world deployment using the Husky robot.

\section*{Declaration of conflicting interests}
The authors declare that they have no known potential conflicts of interest with respect to the research, authorship, and/or publication of this article.

\bibliographystyle{apalike} 
\bibliography{references}

\end{document}